\documentclass[acmlarge]{acmart}

\DeclareUnicodeCharacter{2212}{-}
\AtBeginDocument{%
  \providecommand\BibTeX{{%
    \normalfont B\kern-0.5em{\scshape i\kern-0.25em b}\kern-0.8em\TeX}}}

\setcopyright{acmcopyright}
\copyrightyear{2023}
\acmYear{2023}

\acmJournal{POMACS}
\acmVolume{37}
\acmNumber{4}
\acmArticle{111}
\acmMonth{8}
\usepackage{graphicx}
\usepackage{textcomp}
\usepackage[ruled,linesnumbered]{algorithm2e}
\usepackage{xcolor}
\usepackage{amsmath}
\usepackage{subfigure}
\definecolor[named]{ACMPurple}{cmyk}{0.55,1,0,0.15}
\definecolor[named]{ACMBlue}{cmyk}{1,0.1,0,0.1}
\definecolor[named]{ACMYellow}{cmyk}{0,0.16,1,0}
\definecolor[named]{ACMOrange}{cmyk}{0,0.42,1,0.01}
\definecolor[named]{ACMRed}{cmyk}{0,0.90,0.86,0}
\definecolor[named]{ACMLightBlue}{cmyk}{0.49,0.01,0,0}
\definecolor[named]{ACMGreen}{cmyk}{0.20,0,1,0.19}
\definecolor[named]{ACMPurple}{cmyk}{0.55,1,0,0.15}
\definecolor[named]{ACMDarkBlue}{cmyk}{1,0.58,0,0.21}
\newcommand{\ci}[1]{{\color{ACMPurple} #1}}

\begin{document}

\title{A survey of deep learning optimizers - first and second order methods}

\author{ROHAN V KASHYAP}
\email{rohankashyap@iisc.ac.in}
\orcid{1234-5678-9012}
\affiliation{%
  \institution{Indian Institute of Science}
  \city{Bangalore}
  \state{Karnataka}
  \country{India}
  \postcode{560012}}
\begin{abstract}
Deep Learning optimization involves minimizing a high-dimensional loss function in the weight space which is often perceived as difficult due to its inherent difficulties such as saddle points, local minima, ill-conditioning of the Hessian and limited compute resources. In this paper, we provide a comprehensive review of $14$ standard optimization methods successfully used in deep learning research and a theoretical assessment of the difficulties in numerical optimization from the optimization literature.
\end{abstract}




\maketitle
\section{Introduction}
\begin{figure}[hbt!]
\centering
\begin{subfigure}
  \centering
  \includegraphics[width=.4\linewidth]{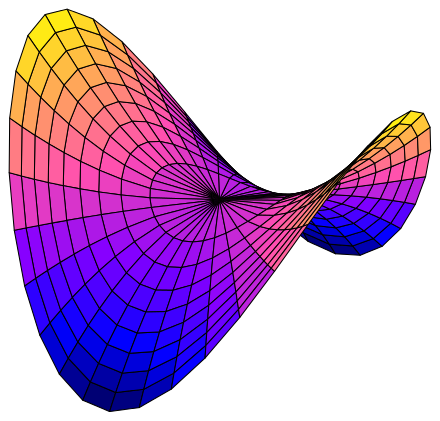}
  \label{fig:sub1}
\end{subfigure}%
\begin{subfigure}
  \centering
  \includegraphics[width=.4\linewidth]{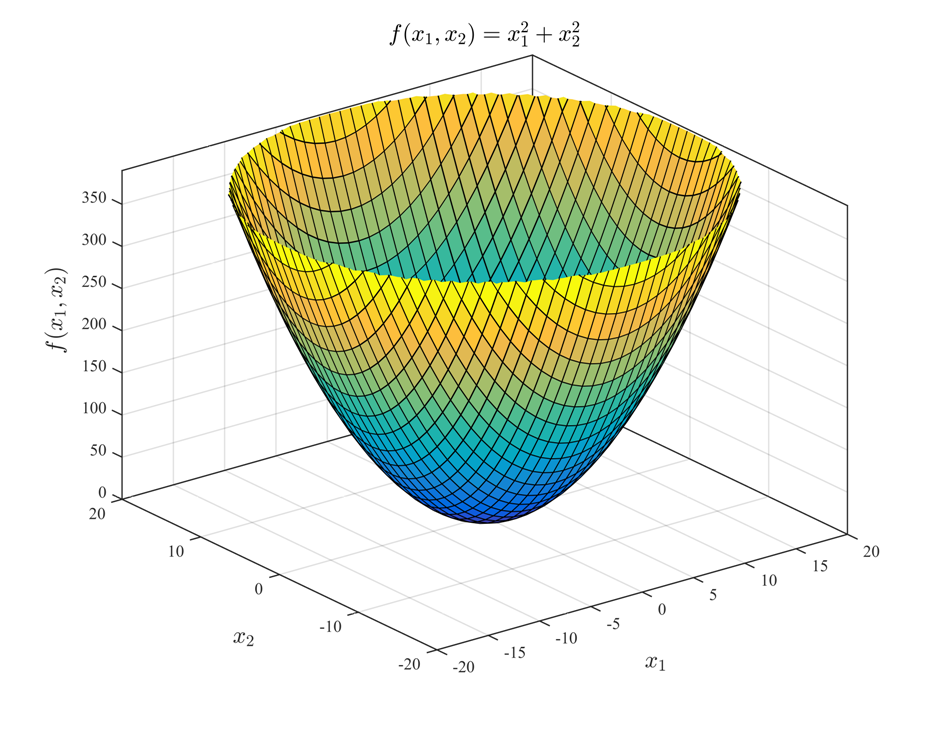}
  \label{fig:sub2}
\end{subfigure}
\end{figure}
\noindent Optimization encompasses a central role in machine learning \cite{b1}, statistical physics, pure mathematics \cite{b11, b16, b30}, random matrix theory and also scientific research. Deep learning involves optimization of non-convex loss functions over continuous, high-dimensional spaces. Neural network training involves solving a loss function that is differentiable and continuous using gradient-based optimization techniques that utilize the first order or second order information to update its weight parameters and thus seek to find the critical point. Gradient descent \cite{b30}, quasi-Newton \cite{b62}, BFGS \cite{b59} and conjugate gradient \cite{b55} methods are commonly used to perform such minimizations and find the optimal solution.\\ 

\noindent \textbf{Definition 1}. Given $K$ weight matrices $(\theta_k)_{k \leq K}$ the output $y$ of a deep neural network with a activation function $\phi$ is given as:
\begin{equation}
    \label{Neural Networks}
f(x) =\phi \left(\phi \left(\cdots \phi \left(x \cdot \theta_1\right) \cdots\right) \cdot \theta_{K-1}\right) \cdot \theta_K,
\end{equation}

\noindent where $x \in R^{d}$ is the input to the model. Let  \( \left\{ x_i, y_i\right\}_{i=1}^m \) denote the set of training examples. In principle, the goal of a deep learning algorithm is to reduce the expected generalization error $\mathbb{E}(L(f(x;\theta), y))$ where $L$ is the designated loss-function such as mean-squared error, and $f(x;\theta)$ is the function estimate for $y = f(x)$ with parameters $\theta$. The most common estimator is the maximum likelihood estimation (MLE) for $\theta$ defined as:
$$
\boldsymbol{\theta}_{\mathrm{ML}}=\underset{\boldsymbol{\theta}}{\arg\max } \hspace{0.2em} \mathbb{E}_{\mathbf{x} \sim \hat{p}_{\text {data }}} \log p_{\text {model }}(\boldsymbol{x} ; \boldsymbol{\theta})
$$
where $p_{\text {model }}$ is the model distribution. This accounts to minimizing the Kullback-Leibler (KL) divergence (a metric) between the empirical data distribution $p_{model}$ and the true data distribution $\hat{p}_{\text {data}}$:
\begin{equation*}
D_{\mathrm{KL}}\left(\hat{p}_{\text {data }} \| p_{\text {model }}\right)=\mathbb{E}_{\mathbf{x} \sim \hat{p}_{\text {data }}}\left[\log \hat{p}_{\text {data }}(\boldsymbol{x})-\log p_{\text {model }}(\boldsymbol{x})\right]
\end{equation*}
Since the true data distribution is largely unknown, we minimize the expected loss on the training set - a quantity called the empirical risk given as follows:
\begin{equation*}
\mathbb{E}_{\boldsymbol{x}, \mathrm{y} \sim \hat{p}_{\mathrm{data}}(\boldsymbol{x}, y)}[L(f(\boldsymbol{x} ; \boldsymbol{\theta}), y)]=\frac{1}{m} \sum_{i=1}^m L\left(f\left(\boldsymbol{x}^{(i)} ; \boldsymbol{\theta}\right), y^{(i)}\right)
\end{equation*}
where $L$ is the loss-function and $f(x;\theta)$ is the prediction output when the input is $x$ and $\hat{p}_{data}$ is the empirical distribution. In a supervised setting, $y$ is the target output and $p(y | x)$ is the probability distribution to be estimated. MLE is the best estimator asymptotically and is consistent in the sense that it converges \footnote{We refer to converges in the sense of pointwise and not in distribution i.e., $\hat{\theta} \rightarrow \theta^*$, as $m \rightarrow \infty$.} to the parameters of the true data distribution $\hat{p}_{\text {data }}$ as $m \rightarrow \infty$. Although this method in theory, is highly prone to overfitting since models with high capacity can simply memorize the training set, empirical results surprisingly enough dictate the opposite and prove the model's capability to achieve local generalization power within its capacity and thus generalizes to unseen examples in the test samples. The solution is to add a penalty term called the regularizer to the empirical risk defined as:
\begin{equation*}
    \Tilde{L}(\boldsymbol{\theta})=\underbrace{\frac{1}{N} \sum_{i=1}^N L(y(\mathbf{x}, \boldsymbol{\theta}), t)}_{\text {training loss }}+\underbrace{\mathcal{R}(\boldsymbol{\theta})}_{\text {regularizer }}
\end{equation*}
where $\mathcal{R}(\boldsymbol{\theta})$ is the regularization term such as $L1$ or $L2$ regularization as illustrated in Fig. \ref{fig:regularization}. However, high-dimensional non-convex optimization techniques come at the cost of no theoretical guarantees yet obtain state-of-the-art results on numerous tasks such as image classification \cite{b64}, text processing \cite{b63}, and representation learning \cite{b65}. Also, deep nets are composed of layers of affine transformations followed by point-wise non-linear activations such as RELU, sigmoid and tanh functions. Thus, the choice of the non-linear function becomes more important to avoid vanishing and exploding gradients problems \cite{b33}.
\subsection{Mathematical Preliminaries and Notations}
\label{Mathematical Preliminaries}
Let $({x^{(i)},y^{(i)}})$ denote the training examples and the corresponding targets respectively. Let $f(.,.;\theta)$ denote a neural network with parameters $\theta$ and a loss function $L$. For $R^n$-valued functions $\nabla$ corresponds to the gradient operator. For all the optimization methods discussed in the paper, $g$ corresponds to the computed gradient vector across multiple training samples, $\epsilon$ is the learning rate and $\rho$ is the decay rate. For first order methods, $r$ is the accumulated square gradient and $v$ is the velocity vector respectively. For second order methods, we denote the Hessian matrix of the loss function $L$ with respect to the model parameters $\theta$ by $H$ and $\lambda_{i}$'s are the corresponding eigenvalues. Unless stated otherwise, the update rule is given as $\theta = \theta -\Delta\theta$, where $\Delta\theta = \epsilon g$. We denote the basic version of stochastic gradient descent algorithm in Algorithm \ref{alg:Alg1}.
\RestyleAlgo{ruled}
\begin{algorithm}[hbt!]
\caption{Stochastic Gradient Descent}
\label{alg:Alg1}
$\textbf{Initialize:}$ Initial parameter $\theta$, Learning rate $\epsilon$.\\
\While{stopping criterion not met}{
    Sample $m$ training examples ${x^{(1)},...,x^{(m)}}$ and their corresponding targets $y^{(i)}s$.\\
    Compute gradient estimate: $g \leftarrow \left[\frac{1}{m} \sum_{i=1}^{m} L\left(f\left(\boldsymbol{x}^{(i)} ; \boldsymbol{\theta}+\alpha \boldsymbol{v}\right), \boldsymbol{y}^{(i)}\right)\right]$\\
    Apply update: $\theta \leftarrow \theta -\epsilon g$.}
\end{algorithm}
\subsection{Literature Overview}
The most important algorithm for machine learning optimization is the stochastic gradient descent method (SGD). This is a general extension to the gradient descent method to handle large training sets without being computationally expensive as discussed in Alg \ref{alg:Alg1}. Since, we perform gradient updates at each step using a mini-batch of examples the asymptotic cost for SGD is $O(1)$ as a function of $m$, where m is the number of training examples. In this section, we provide a detailed discussion of the various problems associated with this method.
\begin{figure}[hbt!]
\centering{\includegraphics[scale=0.3]{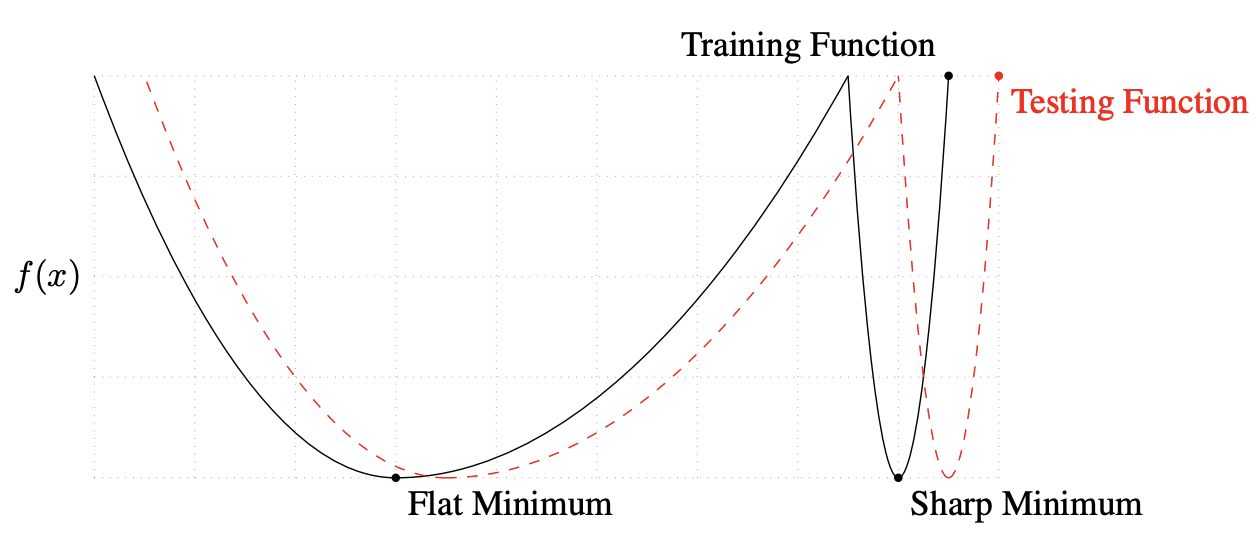}}
    \caption{Illustration of flat and sharp minima. Source Keskar \cite{b46}. }
    \label{fig:galaxy}
    \vspace{-0.5em}
\end{figure}
\subsection{Flat Minima}
\ci{Hochreiter et al.} \cite{b45} conjectured that the generalization of deepnets is related to the curvature of the loss at the converged solution. Their algorithm the "flat minima search" utilizes the Hessian information to find a large region in the weight space called the "flat minima" where all the solutions lead to a small error function value. It is also noted that high error local minima traps do not appear when the model is overparameterized, and \ci{Keskar et al.} \cite{b46} demonstrated that the large batch methods always tend to converge to sharp minimizers (large number of positive eigenvalues) and thus generalize a bit worse than the small batch methods that are known to converge to flat minimizers which are characterized by having numerous small eigenvalues though they have comparable training accuracies \cite{b4}.\\ 

\noindent This is attributed to the empirical evidence as discussed in \cite{b13}, \cite{b45} and \cite{b46} that large batch methods are attracted to regions of sharp minima while the basins found by small batch methods are wider, and since they are unable to escape these basins of attractions, they generalize better. Although a larger batch size provides a more accurate approximation of the true gradient with fewer fluctuations in the update step, it is computationally very expensive and leads to fewer update steps, while a smaller batch size offers a regularizing effect due to the noise they add during the learning process. However, this might lead to high variance in the gradient estimate and thus requires a smaller learning rate to maintain stability and converge to an optimal solution. It turns out that most directions in the weight space have similar loss values since the Hessian consists of a large number of eigenvalues close to zero, even at random initialization points.
\begin{figure}[hbt!]
\centering
\begin{subfigure}
  \centering
  \includegraphics[width=.3\linewidth]{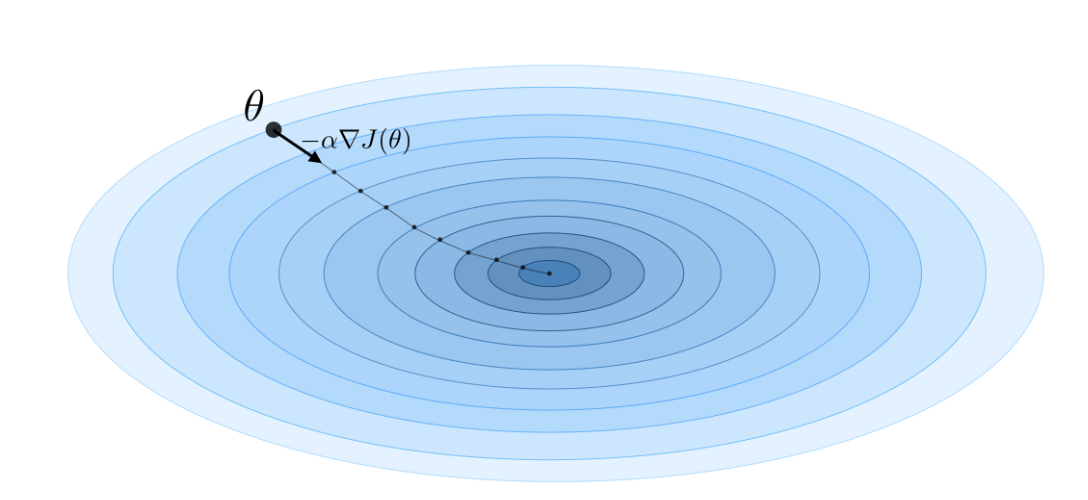}
  \label{fig:sub1}
\end{subfigure}%
\begin{subfigure}
  \centering
  \includegraphics[width=.3\linewidth]{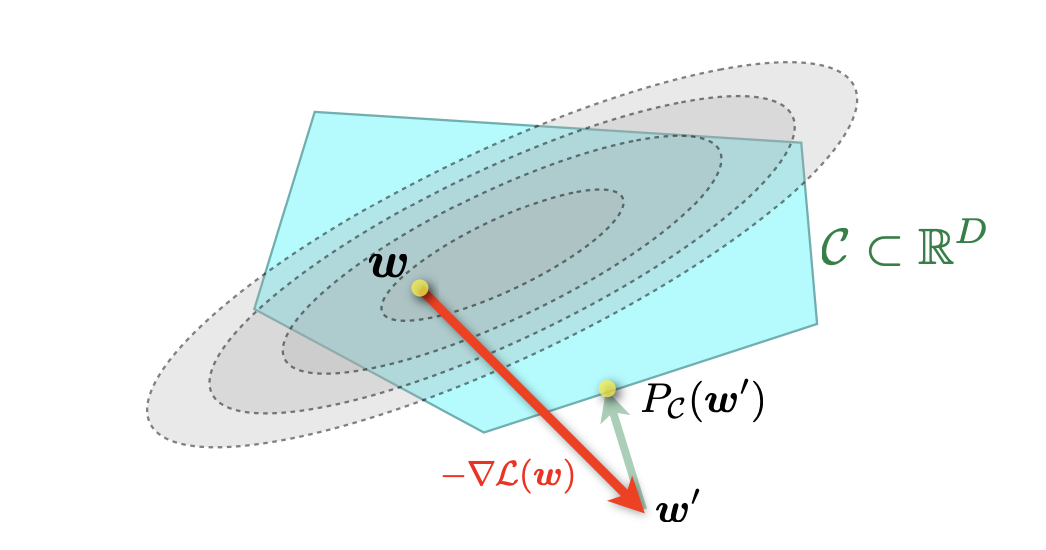}
  \label{fig:sub2}
\end{subfigure}
\caption{a. The $2-D$ visualization of the SGD trajectory. Source: Shervine \cite{b67}. b. Illustration of projected gradient descent where $w$ is the current parameter estimate, $w'$ is the update after a gradient step, and $PC(w')$ projects this onto the constraint set $C$.  Source: Ian Kevin P. Murphy \cite{b44}}
\end{figure}
\subsection{Linear Subspace}
\ci{Goodfellow et al.} \cite{b47} show the existence of a linear subspace during neural network training with no barriers and a smooth path connecting initialization and the final minima. Also, they show that poor conditioning of the Hessian and variance in the gradient estimate are significant obstacles in the SGD training process. These could deviate the algorithm completely from the desired basin of attraction. Thus the final minima or the final region in the parameter space and the width of the final minima that SGD converges to depend largely on the learning rate, batch size, and gradient covariance, with different geometries and generalization properties.
\subsection{SGD Trajectory}
\begin{figure}[hbt!]
\centering{\includegraphics[scale=0.2]{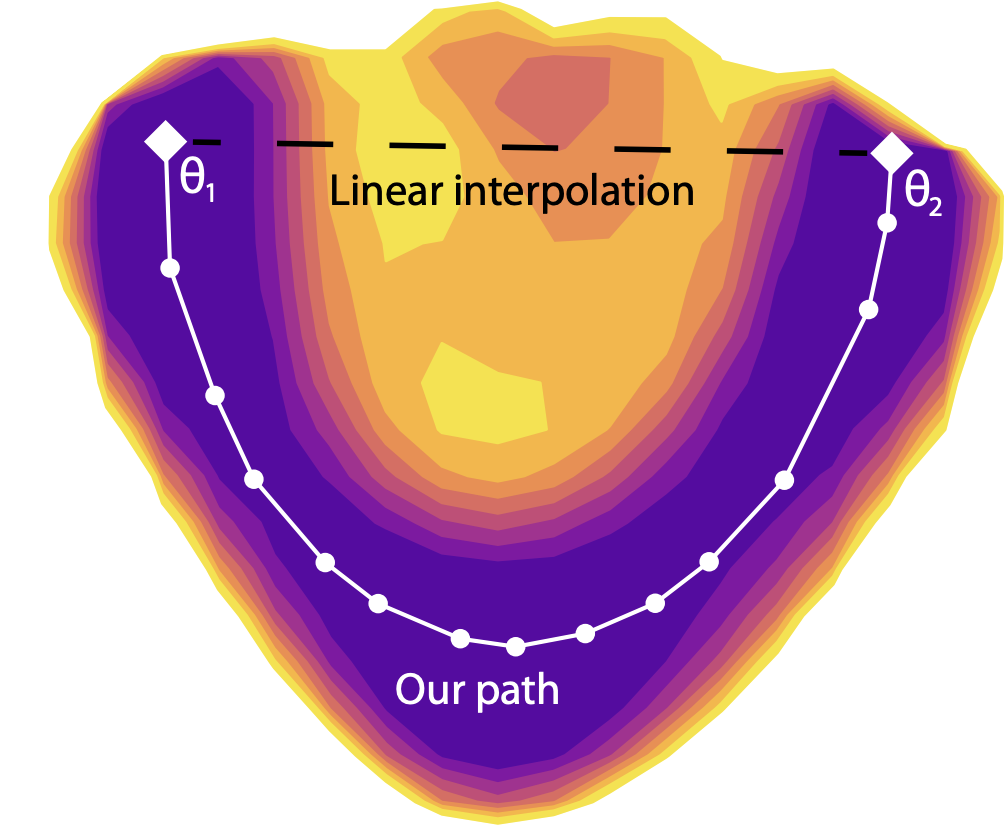}}
    \caption{Left: A slice through the one million-dimensional training loss function of DenseNet-40-12 on CIFAR10 and the minimum energy path as discussed in \cite{b13}. The plane is spanned by the two minima and the mean of the path nodes. Source: Felix Draxler \cite{b13}}
    \label{Figure 1}
    \vspace{-1em}
\end{figure}

\ci{Felix et al.} \cite{b13} constructed continuous paths between minima for recurrent neural networks and conjectured that the loss minima are not isolated points but essentially form a continuous manifold. Also, since these paths are flat, this implies the existence of a single connected manifold of low loss and not a distinct basin of attraction. More precisely, the part of the parameter space when the loss remains below a certain low threshold form a connected region of low error valleys as shown in Figure \ref{Figure 1}. These reflect the generalization capabilities of the network since this provides evidence that large neural nets have enough parameters to produce accurate predictions even though they undergo huge structural changes. This is quite a deviation from the current literature, where the minima are typically depicted as points at the bottom of a strictly convex valley of a certain width with network parameters given by the location of the minimum. 
\subsection{SGD Analysis}
\ci{Stanislaw et al.} \cite{b40} conjectured that the SGD process depends on the ratio of the learning rate to batch size, which thus determines the width of the endpoint and its generalization capabilities. First, as we approach the local minima, the loss surface can be approximated by a quadratic bowl, with the minimum at zero loss. Thus, the training can be approximated by an Ornstein-Unhlenbeck \footnote{The Ornstein-Unhlenbeck $x_t$ process is a Gaussian process defined using the following differential equation: $d x_t=-\theta x_t d t+\sigma d W_t$.} process as discussed in Section \ref{Second-order methods}. Second, the covariance of the gradients $C=\frac{1}{N} \sum_{i=1}^N\left(g_i-g\right)^T\left(g_i-g\right)$, where $g_{i}$ is the gradient of the loss function $L$ with respect to the ($x_{i}, y_{i}$) training example, and the Hessian of the loss function $H$ are approximately equal, and is possibly the reason SGD escapes the global minima.
\noindent Since the loss functions of deep neural networks are typically non-convex, with complex structure and potentially multiple minima and saddle points, SGD generally converges to different regions of parameter space, with different geometries and generalization properties \cite{b2, b3}, depending on optimization hyper-parameters and initialization.

\subsection{Curse of Dimensionality}
Finding a good descent direction in a high-dimensional space is a difficult problem, but it is not nearly as difficult as navigating an error surface with complicated obstacles within multiple low-dimensional subspaces. It is also highly likely that the non-convex loss landscape is such that it contains regions of local minima at high energies which full-batch methods couldn't have possibly avoided. But, the stochastic nature of SGD due to its inherent noise and a smaller batch size doesn't suffer from this problem and is not trapped in irrelevant basins of attraction.
\noindent With non-convex functions, such as neural nets, it is possible to have many local minima. Indeed, nearly any deep model is guaranteed to have an extremely large number of local minima. However, this is not necessarily a major problem. It is empirically observed that as the dimensionality $N$ increases, local minima with high error relative to the global minimum occur with a probability that is exponentially small in $N$ \cite{b5}. This is mainly because neural nets exhibit model identifiability problem, which results in an extremely large amount of local minima, but all are equivalent in their cost function value. They do not have very high error values relative to the global minima and are thus good solutions that stochastic gradient descent often finds itself at during the training process.
\subsection{Critical Points}
Critical points \footnote{A critical point is a point whose first-order derivative is zero i.e., $f'(x) = 0$.} with high costs are far more likely to be saddle points \cite{b5}. We are not exactly concerned with finding the exact minimum of a function but seek only to reduce its value significantly to obtain a good generalization error. In the gradient descent method, the main problem is not the direction but its step size along each eigenvector because the update step is directly proportional to the corresponding eigenvalue. Thus, we move towards the optimal point if the eigenvalue is positive else; we move away from it if it is negative along those directions, and this slows down the learning process since now we might circumnavigate a tall mountain or a flat region of high error or might simply take a very long time to move away from the saddle points.\\
\begin{figure}[hbt!]
\centering{\includegraphics[scale=0.3]{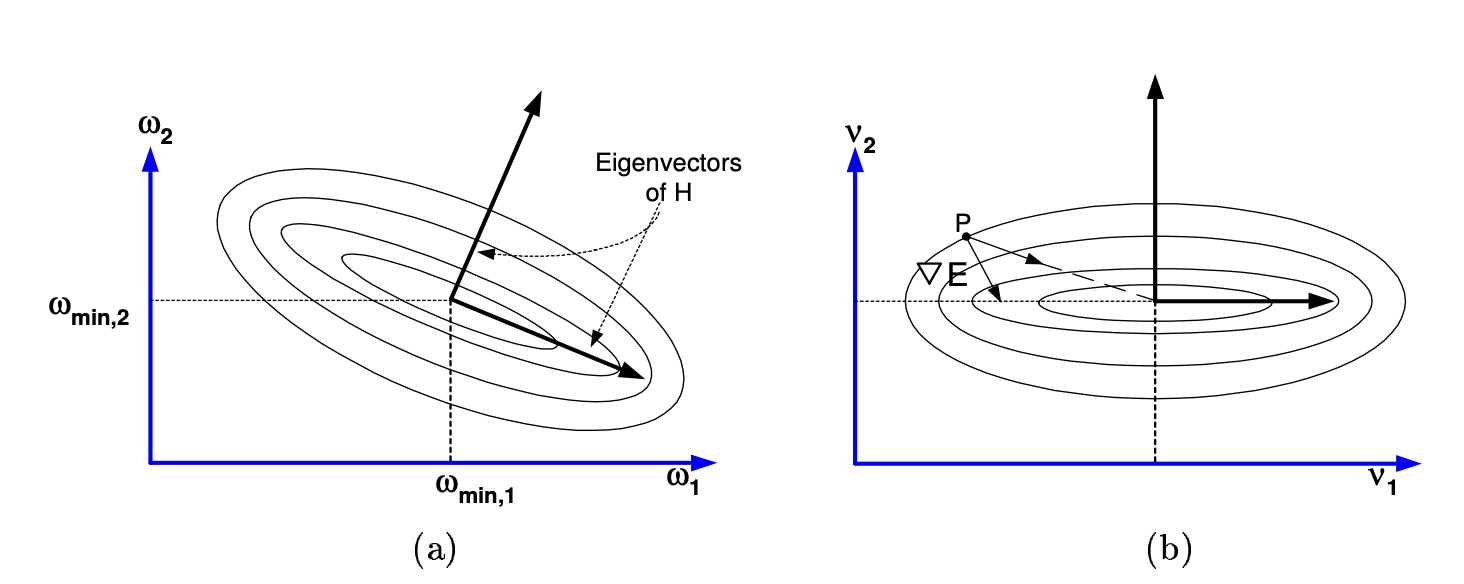}}
    \caption{For two dimensions, the lines of constant of the error surface E are oval in shape. The eigenvectors of H along the major and minor axes. The eigenvalues measure the steepness of E along each eigendirection. Source:Yann LeCun \cite{b1}}
    \label{fig:galaxy}
\end{figure}

\noindent For a convex function, such a flat region corresponds to the global minima, but in a non-convex optimization problem, it could correspond to a point which a high value of the loss function. Although, Newton's method solves this slowness issue by rescaling the gradients in each direction with the inverse of Hessian, it can often take the wrong direction and be applied iteratively as long as the Hessian remains positive definite. Thus for a locally quadratic function, Newton's method jumps directly to the minimum, but if the eigenvalues are not all positive (near a saddle point), then the update can lead us to move in the wrong direction. Also, since second-order methods are computationally intensive, the computational complexity of computing the Hessian is $ O\left(N^{3}\right) $, and thus, practically only networks with a small number of parameters can be trained via Newton's method.\\

\noindent Several methods such as BFGS and saddle-free newton method \cite{b5, b53, b54} are proposed to overcome this problem in non-convex settings, specifically when the Hessian has negative eigenvalues. This includes adding a constant along the Hessian diagonal, which largely offsets the negative eigenvalues. Recently, Luke \cite{b48} established the connections between gradient explosion and the initialization point by measuring the recurrent Jacobian. They found that with a  random initialization, the neural network's policy is poorly behaved and has many eigenvalues with a norm greater than length $1$, and thus the gradient norm grows. However, for a stable initialization (shrinks the initialization by multiplying by $0.01$) which was picked specifically to prevent the gradients from exploding, we find that many eigenvalues close to $1$, and thus the gradients do not explode. \ci{Sutskever et al.} \cite{b9} demonstrated that poorly initialized and the absence of momentum perform worse than well-initialized networks with momentum.
\begin{figure}[hbt!]
\centering{\includegraphics[scale=0.25]{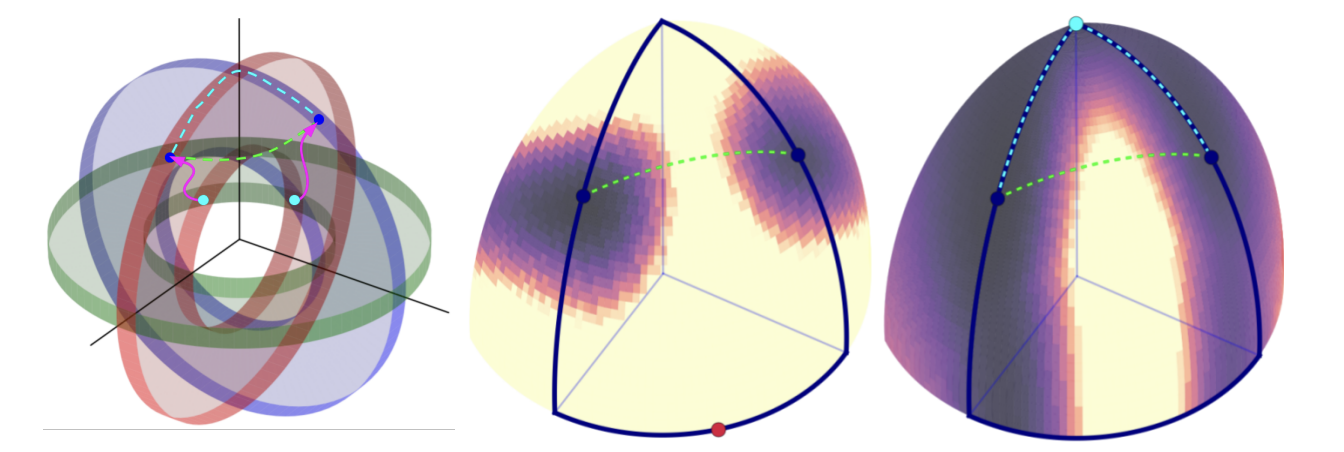}}
    \caption{The loss surface represented as a union of $n$-dimensional manifolds called $n$-wedges. A model of the low-loss manifold comprising of $2$-wedges in a $3$-dimensional space. Source: Stanislav \cite{b8}}
    \label{fig:quadratic approximation}
\end{figure}

\subsection{Difficulties in Neural Network Optimization}
The main challenges for neural net optimization in these high-dimensional spaces can be ill-conditioning of the Hessian, poor condition number, local minima, plateaus, and flat regions, the proliferation of saddle points, and poor correspondence between local and global structure. The existence of a large number of saddle points can significantly slow down learning since they are regions of high error plateaus and give the impression of the presence of a local minimum. Also, since these critical points are surrounded by plateaus of small curvature, second-order methods such as the Newton method are attracted to saddle points and thus slow down the learning process \cite{b5}. Also, a common observation is that at the end of training process, the norm of the gradients $\nabla f(x)$ is not necessarily small i.e., $\nabla f(x) < \epsilon$, and the Hessian $H$ has negative eigenvalues, indicating that the algorithm did not converge to the local minima \cite{b23}.\\

\noindent It is also possible that models with high capacity have a large number of local minima which can increase the number of interactions and cause the Hessian to be ill-conditioned. There are also wide, flat regions of constant value where the gradients and Hessian are all zero. Such degenerate locations pose a significant problem to optimization algorithms. We conjecture that large neural nets often converge to local minima, most of which are equivalent and yield similar generalization performance on the test set. Also, the probability of finding a high error local minima is exponentially small in high-dimensional structures and decreases quickly with network size.
\begin{figure}[hbt!]
\centering{\includegraphics[scale=0.2]{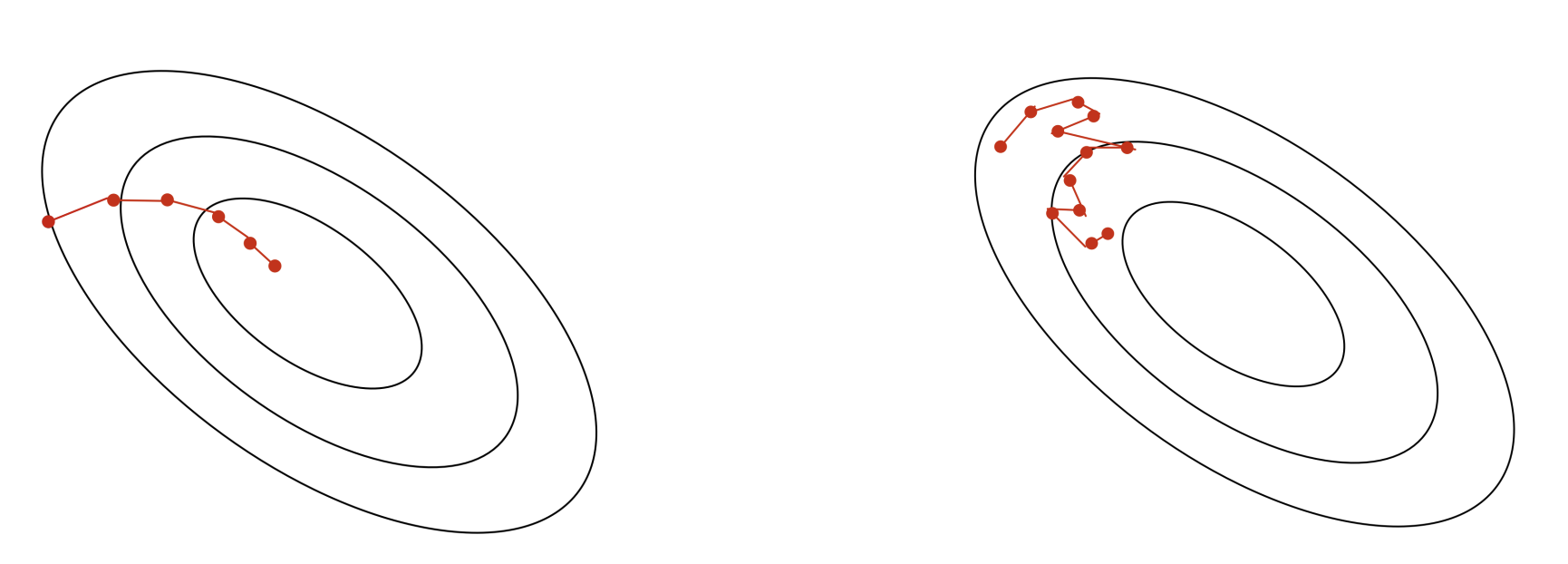}}
    \caption{a. Batch gradient descent moves directly downhill. b. SGD takes steps in a noisy direction, but moves downhill on average. Source: Roger Grosse \cite{b21} }
    \label{fig:galaxy}
    \vspace{-1em}
\end{figure}
\section{First Order Methods}
\subsection{Momentum}
\label{momentum}
\ci{Polyak et al.} \cite{b49} introduced the momentum method to accelerate the learning process of first-order gradient-based methods, especially in regions of high curvature. It also addresses the poor conditioning of the Hessian matrix and the variance in the stochastic gradient descent process. This algorithm computes an update step by accumulating an exponentially decaying moving average of the past gradients and continues to move in their direction. The momentum term increases for dimensions whose gradients point in the same directions and reduces updates for dimensions whose gradient change directions, and we gain a faster convergence. Although it can be applied to both full-batch and mini-batch learning methods, stochastic gradient descent combined with mini-batches and a momentum term has been the golden rule in the deep learning community to get excellent results on perception tasks. It dampens oscillations in directions of high curvature by combining gradients of opposite signs and also yields a larger learning rate in regions of low curvature. It builds but the speed in directions of consistent gradient and thus traverses a lot faster than the steepest descent due to its accumulated velocity.\\

\noindent If the momentum is close to $1$, this algorithm is a lot faster, and it is equally essential to ensure a smaller momentum value (close to $0.5$) at the start of the learning process. In the beginning, since the points are randomly initialized in the weight space, there may be large gradients, and the velocity term ($v$) can lead in not being entirely beneficial. Thus, once these large gradients disappear and we find the desired basin of attraction, we can increase the momentum term smoothly to its final value, which is $0.9$ to $0.99$, to ensure that learning is not stuck and wastes computational time in traversing flat regions where the gradients are almost zero.
$$
\begin{array}{l}
\boldsymbol{v} \leftarrow \alpha \boldsymbol{v}-\epsilon \nabla_{\theta}\left(\frac{1}{m} \sum_{i=1}^{m} L\left(\boldsymbol{f}\left(\boldsymbol{x}^{(i)} ; \boldsymbol{\theta}\right), \boldsymbol{y}^{(i)}\right)\right) \\
\boldsymbol{\theta} \leftarrow \boldsymbol{\theta}+\boldsymbol{v}
\end{array}
$$
\noindent The momentum algorithm accelerated convergence to a local minimum, requiring fewer iterations than the steepest descent method, precisely by $ \sqrt{R} $ times fewer iterations, where R is the condition number of the curvature at the local minimum. The velocity term plays the role of velocity as in physical analogy in which it accelerates the particle through the parameter space in reinforced directions, and thus the velocity vector is also known as imparting momentum to the particle. The hyperparameter determines how quickly the contributions of previous gradients exponentially decay. Previously in gradient descent, the step size was proportional to the norm of the gradient. Now, it depends on how large and aligned a sequence of gradients are. The step size is largest when many gradients point in exactly the same direction. We can think of the analogy of a ball rolling down a curved hill, and whenever it descends a steep part of the surface, it gathers speed and continues sliding in that direction.

\subsection{Nesterov Momentum}
\begin{figure}[hbt!]
\centering{\includegraphics[scale=0.25]{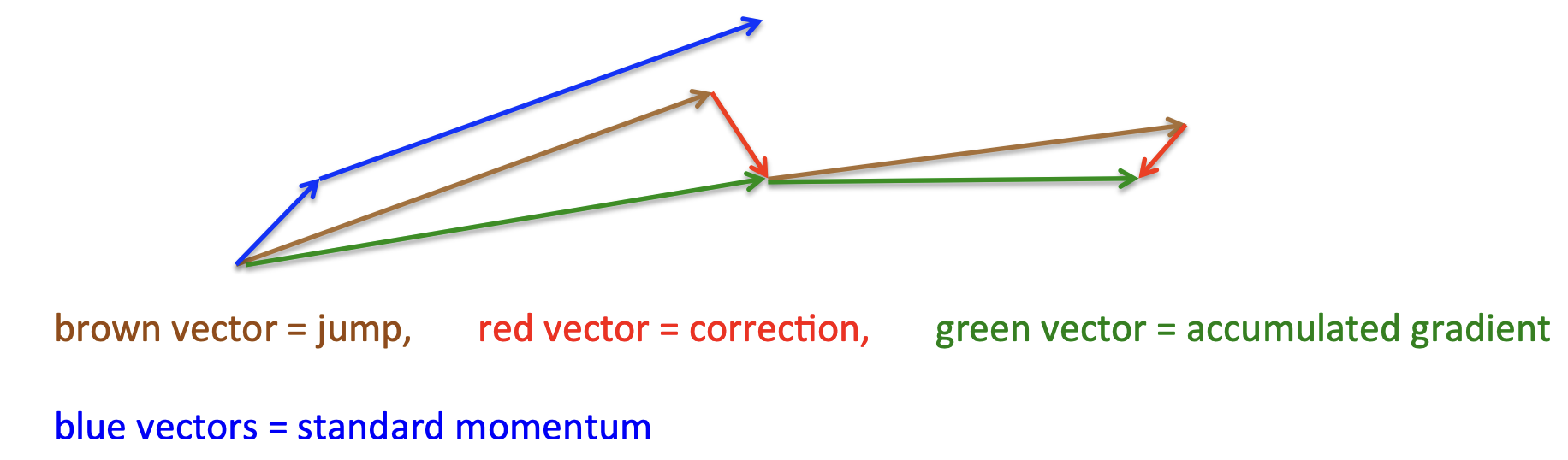}}
    \caption{Nesterov momentum first makes a big jump in the direction of the previously accumulated gradient and then computes the gradient where you end up and thus makes a correction. Source: Geoffrey Hinton \cite{b29}}
    \label{fig:galaxy}
    \vspace{-1em}
\end{figure}
\noindent Sutskever \cite{b9} introduced a variant of the classical momentum (CM) that aligns in line with the work of Nesterov’s accelerated gradient (NAG) method for optimizing convex functions. The update rule is given as:
\begin{equation*}
\boldsymbol{v} \leftarrow \alpha \boldsymbol{v}-\epsilon \nabla_{\boldsymbol{\theta}}\left[\frac{1}{m} \sum_{i=1}^{m} L\left(f\left(\boldsymbol{x}^{(i)} ; \boldsymbol{\theta}+\alpha \boldsymbol{v}\right), \boldsymbol{y}^{(i)}\right)\right]
\end{equation*}
\begin{equation*}
\boldsymbol{\theta} \leftarrow \boldsymbol{\theta}+\boldsymbol{v}
\end{equation*}
\noindent where $\alpha \in [0, 1]$ is a hyperparameter that determines the contributions of previous gradients across multiple iterations. From the formula, it is clear that NAG is similar to CM\footnote{CM is the abbreviation for Classical Momentum as discussed in Section \ref{momentum}.} except where the gradient is evaluated. With Nesterov momentum, the gradient is evaluated after the current velocity is applied. This means that in NAG we first make a big jump in the direction of the previously accumulated gradient and then compute the gradient where you end up and thus make a correction. This gradient-based correction factor is vital for a stable gradient update, especially for higher values of µ. The gradient correction to the velocity responds much quicker in NAG, and if $ \mu v_{t} $ is indeed a poor update and an inappropriate velocity, then NAG will point back towards $\theta_{t}$ more strongly than $ \nabla f\left(\theta_{t}\right) $ does, thus providing a larger and more timely correction to $ v_{t} $ than CM. Therefore, it can avoid oscillations and is much more effective than CM along high-curvature vertical directions. Thus, it is more tolerant to larger values of µ than CM. For convex functions, Nesterov momentum achieves a convergence rate of $ O\left(1 / K^{2}\right) $ where k is the number of steps.
\begin{figure}[hbt!]
\centering
\begin{subfigure}
  \centering
  \includegraphics[width=.25\linewidth]{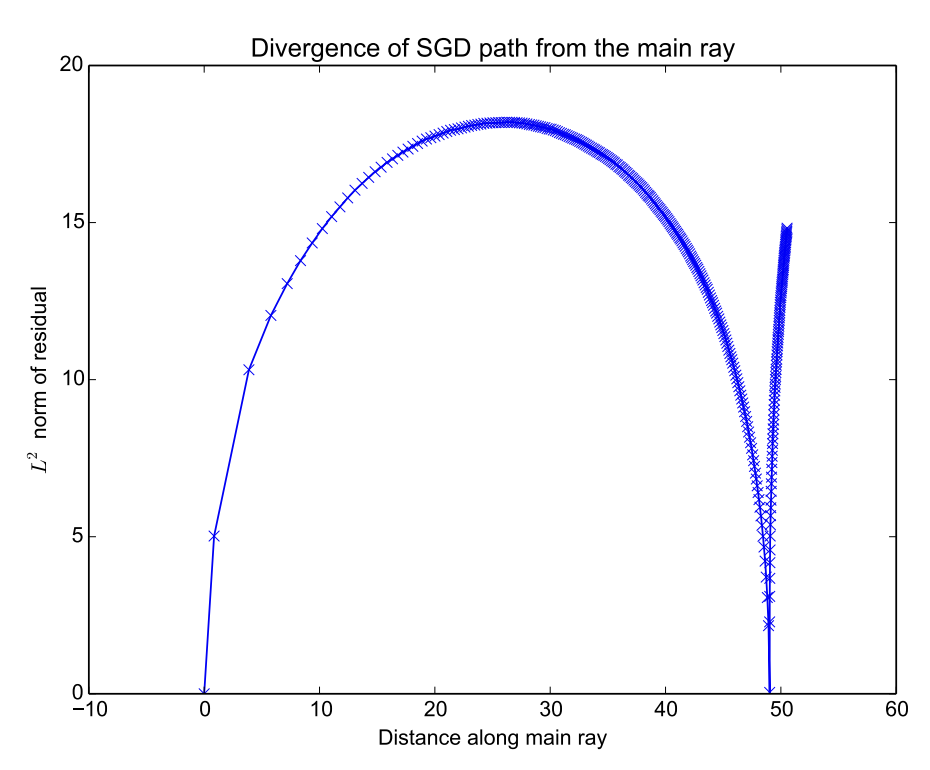}
  \label{fig:sub1}
\end{subfigure}%
\begin{subfigure}
  \centering
  \includegraphics[width=.25\linewidth]{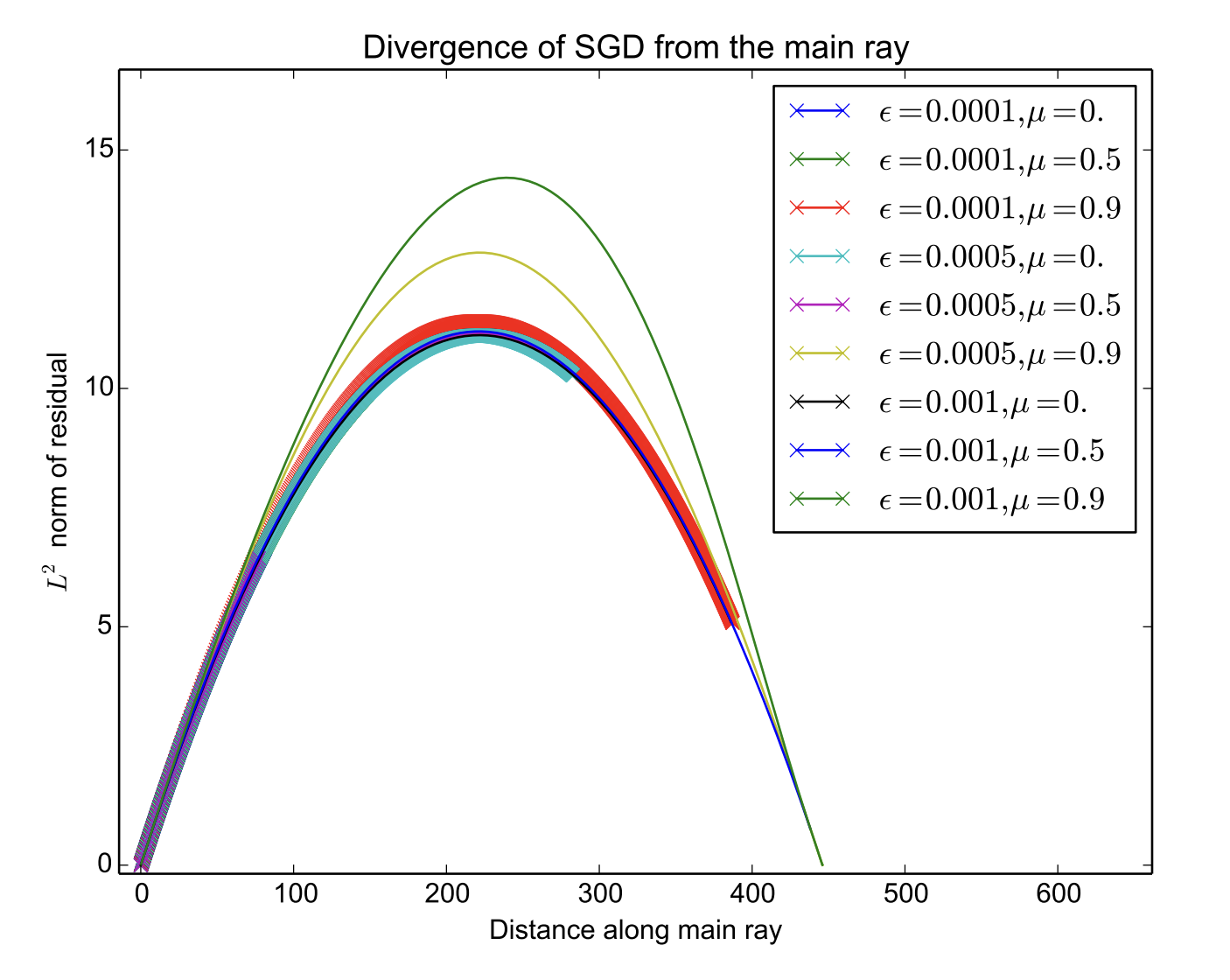}
  \label{fig:sub2}
\end{subfigure}
\caption{The effect of different learning rate $\epsilon$ and the momentum coefficients $\mu$ for various hyperparameters. (left) The plot represents the norm of the residual of the parameter value after projecting the parameters into a $1$-D subspace $\forall t$. (right) The divergence of SGD from the main linear path when trained using a maxout network on MNIST. Source: Goodfellow \cite{b35}}
\end{figure}
\subsection{Adaptive Learning Rates}
The delta-bar-delta algorithm \cite{b50} introduced the concept of having separate adaptive learning rates for each individual connection (weight) which is set empirically by observing its gradient after each iteration. The idea is that if the gradient stays consistent, i.e., remains the same sign, we tune up the learning rate; else if the gradient sign reverses, we decrease its learning rate. The intuition is that for deep neural nets, the appropriate learning rate varies widely for each of its weights. The magnitude of the gradients is often different for different layers, and if the weight initialization is small then the gradients are small for early layers than the later ones for very deep neural nets. We start with a local gain of $1$ for each weight and use small additive increases or multiplicative decreases depending on the sign of the gradient (mini-batch). This ensures that the big gains decay rapidly when the oscillations start. It is essential to ensure that we limit the gains to lie within a reasonable range and use bigger mini-batches to ensure that the sign of the gradient is not due to the sampling error of the mini-batches.\\
Jacobs \cite{b50} proposed a method for combining adaptive learning rates with momentum. He conjectures that instead of using the sign agreement between the current gradient and the previous gradient, we use the current gradient and the velocity of the weight and thus combine the advantages of momentum and adaptive learning rates. Since momentum doesn't care about axis-aligned effects, it can deal with diagonal ellipses and traverse in diagonal directions quickly.
\subsection{Adagrad}
The Adagrad \cite{b51} algorithm individually adapts the learning rates of all model parameters by scaling them inversely proportional to the square root of the sum of the historical squared values of the gradient.
\begin{equation*}
    \boldsymbol{g} \leftarrow \frac{1}{m} \nabla_{\boldsymbol{\theta}} \sum_i L\left(f\left(\boldsymbol{x}^{(i)} ; \boldsymbol{\theta}\right), \boldsymbol{y}^{(i)}\right)
\end{equation*}
\begin{equation*}
    \boldsymbol{r} \leftarrow \boldsymbol{r}+\boldsymbol{g} \odot \boldsymbol{g}
\end{equation*}
For weight parameter with the largest gradient has the largest decrease in its learning rate, while the parameter with the smallest gradient has the smallest decrease in its learning rate. This results in faster progress in the more gently sloped regions of parameter space and works well when the gradient is sparse. For non-convex optimization problems, accumulating the squared gradients can result in a premature and excessive decrease in the learning rate. The parameter update is given as:
\begin{equation*}
    \Delta \boldsymbol{\theta}=-\frac{\epsilon}{\sqrt{\delta+\boldsymbol{r}}} \odot \boldsymbol{g}
\end{equation*}

\subsection{RMSProp}
The RMSProp algorithm \cite{b29} modifies the Adagrad by changing the accumulated squared gradient into an exponentially weighted moving average. In the non-convex setting, since Adagrad decreases the learning rate rapidly by using the history of the squared gradient, it is possible that the learning halts before arriving at the desired basin of attraction or locally convex bowl structure. RMSProp uses an exponentially decaying average of the squared gradients to discard the past history and can converge rapidly after finding a convex bowl it behaves like an instance of the AdaGrad algorithm since the AdaGrad method is designed to converge rapidly for a convex function.
\begin{equation*}
    \boldsymbol{r} \leftarrow \rho \boldsymbol{r}+(1-\rho) \boldsymbol{g} \odot \boldsymbol{g}
    \vspace{0.5em}
\end{equation*}
If the eigenvectors of the Hessian are axis-aligned, then RMSprop can correct the curvature. Since RMSProp lacks the bias-correction term, it can often lead to large step sizes and divergence with sparse gradients. If the eigenvectors of the Hessian are axis-aligned (dubious assumption), then RMSProp can correct the curvature.

\subsection{Adam}
Adam \cite{b18} (adaptive moments) aligns along the works of RMSProp \cite {b29} and momentum with a few distinctions, such as invariance to the diagonal rescaling of the gradients. First, in Adam, we incorporate momentum by computing the biased first-order moment of the gradient as an exponentially decaying moving average. Second, similar to the RMSProp algorithm, we incorporate the second-order moment of the gradient as an exponentially decaying average. This combination has theoretical guarantees in convex settings. Since the moving averages are initialized to a vector of all zeros, the moment estimates are biased during zero during the initial steps especially when the biased first moment ($s$) and and second moment ($r$) estimates are close to $1$. \footnote{$\odot$ corresponds to element-wise product.}
\begin{equation*}
   \boldsymbol{s} \leftarrow \rho_1 \boldsymbol{s}+\left(1-\rho_1\right) \boldsymbol{g};\hspace{1em}
   \boldsymbol{r} \leftarrow \rho_2 \boldsymbol{r}+\left(1-\rho_2\right) \boldsymbol{g} 
   \vspace{0.5em}
\end{equation*}
Thus, we introduce bias corrections to these estimates to account for their initialization for both the momentum term and the second-order moment (uncentered variance). 
\begin{equation*}
    \hat{\boldsymbol{s}} \leftarrow \frac{\boldsymbol{s}}{1-\rho_1^t};\hspace{1em}
    \hat{\boldsymbol{r}} \leftarrow \frac{\boldsymbol{r}}{1-\rho_r^t}
\end{equation*}
Since RMSProp lacks the correction factor, it has a high-bias in the early stages of training, while Adam is robust to the choice of hyperparameters. Since the accumulated squared values of the gradients term are an approximation to the diagonal of the Fisher information matrix, it is more adaptive and leads to faster convergence. The parameter update is given as:\footnote{In Adam optimization, $\delta$ corresponds to a small constant of the order of $10^{-6}$ to ensure numerical stability.} 
\begin{equation*}
  \Delta \boldsymbol{\theta}=-\epsilon \frac{\hat{\boldsymbol{s}}}{\sqrt{\hat{\boldsymbol{r}}}+\delta}  
\end{equation*}
\section{Second-order methods}
\begin{figure}[hbt!]
\centering{\includegraphics[scale=0.2]{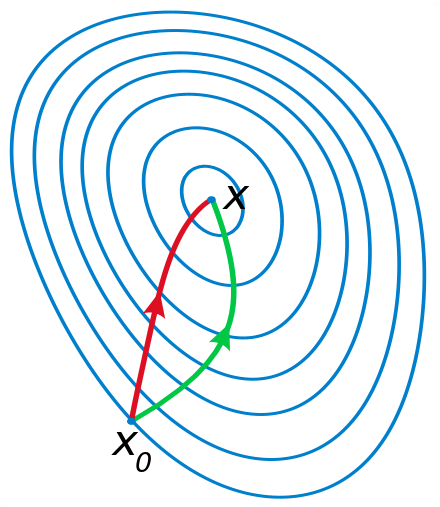}}
    \caption{Illustration of the gradient descent and Newton’s method for a quadratic function. Newton’s method converges in a single step (direct route) by exploiting the curvature information contained in the Hessian matrix.}
\label{fig:galaxy}
\vspace{-1em}
\end{figure}
\subsection{Preliminaries and Motivation}
\begin{figure}[hbt!]
\centering{\includegraphics[scale=0.4]{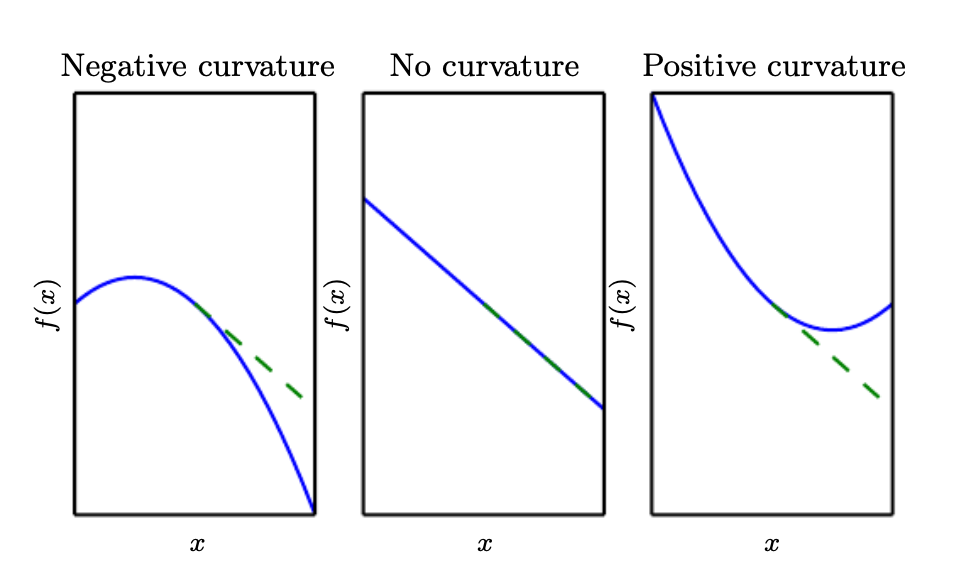}}
    \caption{A quadratic function with various curvature. The dashed line indicates the expected descent direction and the value of the cost function downhill performed by gradient descent. With negative curvature, the cost function decreases faster than the gradient predicts, while with positive curvature, the function decreases much more slowly than expected and thus begins to increase. With no curvature, the gradient correctly predicts the decrease. Source: Ian Goodfellow \cite{b23}}
    \label{fig:quadratic function}
\end{figure}

\noindent For a scalar-valued function, the matrix containing the second-order derivatives is the Hessian matrix ($H$). The Hessian $H$ is a measure of curvature or concavity of the function as shown in Figure \ref{fig:quadratic function}. The matrix containing all the first-order derivatives for vector-valued functions $ f: R^n \rightarrow R^m$ is known as the Jacobian matrix ($J$). Suppose $ f: R^{n} \rightarrow R$ (input is a vector and output is scalar), then the Hessian matrix $H$ and the Jacobian $J$ for $f$ is given as:
$$
\mathbf{J}=
\begin{bmatrix}
\nabla^{\mathrm{T}} f_{1} \\
\vdots \\
\nabla^{\mathrm{T}} f_{m}
\end{bmatrix}
=
\begin{bmatrix}
\frac{\partial f_{1}}{\partial x_{1}} & \cdots & \frac{\partial f_{1}}{\partial x_{n}} \\
\vdots & \ddots & \vdots \\
\frac{\partial f_{m}}{\partial x_{1}} & \cdots & \frac{\partial f_{m}}{\partial x_{n}}
\end{bmatrix}
\quad
\mathbf{H}=
\begin{bmatrix}
\frac{\partial^{2} f}{\partial x_{1}^{2}} & \frac{\partial^{2} f}{\partial x_{1} \partial x_{2}} & \cdots & \frac{\partial^{2} f}{\partial x_{1} \partial x_{n}} \\
\frac{\partial^{2} f}{\partial x_{2} \partial x_{1}} & \frac{\partial^{2} f}{\partial x_{2}^{2}} & \cdots & \frac{\partial^{2} f}{\partial x_{2} \partial x_{n}} \\
\vdots & \vdots & \ddots & \vdots \\
\frac{\partial^{2} f}{\partial x_{n} \partial x_{1}} & \frac{\partial^{2} f}{\partial x_{n} \partial x_{2}} & \cdots & \frac{\partial^{2} f}{\partial x_{n}^{2}}
\end{bmatrix}
\vspace{1em}
$$
\noindent The Hessian matrix $H$ is real and symmetric with $H_{i j}=H_{j i}$. Likewise, the condition number of the Hessian is defined as the ratio of the largest eigenvalue to the smallest eigenvalue, i.e., is the ratio of the steepest ridge's steepness to the shallowest ridge's steepness. Since the Hessian is symmetric with real eigenvalues, and the eigenvectors have an orthogonal basis its eigendecomposition of the matrix is given as:
\begin{equation*}
    \label{decomposition}
   \boldsymbol{H}=\boldsymbol{Q} \operatorname{diag}(\boldsymbol{\lambda}) \boldsymbol{Q}^{-1} 
\end{equation*}
\noindent where $Q$ is an orthogonal matrix ($Q^{T} Q=I$) with one eigenvector per column and $\lambda$ is a diagonal matrix. In higher dimensions, the eigendecomposition of the Hessian can be used to test the critical point. The point is a local minimum when $H$ is positive definite (all $\lambda_{i} > 0$). Likewise, when $H$ is negative definite (all $\lambda_{i}$ < 0), the point is a local maximum. A saddle point has both positive and negative eigenvalues, where it is a local minima across one cross-section and a local maxima within another cross-section.\\

\noindent Second-order optimization methods \cite{b30,b34, b52} utilize the second derivatives for weight update. Newton's method arises naturally from the second-order Taylor series approximation of the function $f$, ignoring the higher-order derivatives is given as: 
$$
f(x) \approx f(x_0) + (x-x_0^{\top}) \nabla_{x} f(x_0)+\frac{1}{2}(x-x_0)^{\top} H(x-x_0)
$$
where $H$ is the Hessian of $f$ with respect to $x$ evaluated at $x_0$. Solving for the critical point $x^*$ we get the Newton update rule:
$$
\label{Newton Update}
x^*=x_0 -H(f) (x_0)^{-1} \nabla_{x} f(x_0)
$$

\noindent Convex functions have strong theoretical guarantees because they are well-behaved and have nice properties. They lack saddle points, and all their local minima correspond to the global minima specifically because the Hessian $H$ is positive semi-definite. For a quadratic function, the Hessian $H$ is constant and thus well-conditioned. A detailed analysis for a quadratic convex function is discussed in Section \ref{Quadratic Function Optimization}. However,  for non-convex functions, the Hessian matrix $\nabla^2 f_k$ may not always be positive-definite, eq. \ref{Newton Update} may not be a descent direction. Also, Hessian being ill-conditioned matters more near the minima since if the Hessian changes too fast we may follow a zig-zag path leading to sub-optimal performance.\\

\noindent To understand the ill-conditioned problem, we consider a convex quadratic objective given as:
\begin{equation*}
\mathcal{J}(\theta)=\frac{1}{2} \theta^{T} A \theta
\end{equation*}
\noindent where $A$ is a positive semi-definite symmetric matrix. If the Hessian of the objective i.e., $\nabla^2 \mathcal{J}(\theta) = A$ is ill-conditioned then gradient also changing rapidly and the update using gradient descent is as follows:\\

$\begin{array}{rlr}\theta_{k+1} & \leftarrow \theta_{k}-\alpha \nabla \mathcal{J}\left(\theta_{k}\right) & \\ & =\theta_{k}-\alpha A \theta_{k} & \\ & =(I-\alpha A) \theta_{k} & \\ \Longrightarrow \theta_{k} & =(I-\alpha A)^{k} \theta_{0} & \\ & =\left(I-\alpha Q \Lambda Q^{T}\right)^{k} \theta_{0} & \\ & =\left[Q(I-\alpha \Lambda) Q^{T}\right]^{k} \theta_{0} & \\ & =Q(I-\alpha \Lambda)^{k} Q^{T} \theta_{0} & \end{array}$\\

\noindent where $Q\Lambda Q^T$ is the eigendecomposition of $A$ as discussed in eq. \ref{decomposition}. The stability conditions is as follows:\\
1. $0<\alpha \lambda_{i} \leq 1:$ decays to $0$ at a rate that depends on $\alpha \lambda_{i}$.\\
2. $1<\alpha \lambda_{i} \leq 2:$ oscillates.\\
3. $\alpha \lambda_{i}>2:$ unstable (diverges).\\

\noindent where $\alpha$ is the learning-rate of the algorithm and $\lambda_{i}$ are the eigenvalues. Hence, we need the learning rate to be bound by $ \alpha<2 / \lambda_{\max } $ to prevent instability. This ensures that we avoid overshooting the minima and not going uphill in directions with strong positive curvature. This also bounds the rate of progress in other directions given as $ \alpha \lambda_{i}<2 \lambda_{i} / \lambda_{\max }$.\\

\noindent Given a starting point $x_i$, one could construct a local quadratic approximation to the objective function ($\mathcal{J}(q)$) that matches the first and second derivative values at that point. We then minimize the approximate (quadratic function) instead of the original objective function. The minimizer of the approximate function is used as the starting point in the next step, and repeat the procedure iteratively. This ensures that the search direction accounts for the curvature information along almost all directions, for all iterations as shown in Figure \ref{fig:quadratic approximation}.\\

\begin{figure}[hbt!]
\centering{\includegraphics[scale=0.4]{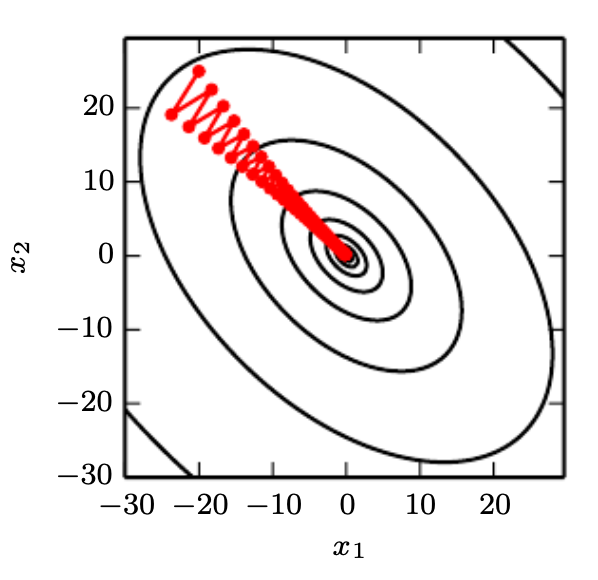}}
    \caption{The first-order gradient descent method fails to exploit the curvature information contained in the Hessian matrix. The condition number greatly influences the SGD trajectory, and we illustrate it using a quadratic function f(x) whose Hessian has a condition number 5. This means that the direction of the most curvature has five times more curvature than that of the least curvature. In this case, the most curvature is in the direction $[1, 1]$, and the least curvature is in the direction $[1, −1]$. The red lines indicate the path followed by gradient descent. This results in an ellipsoidal quadratic function, and gradient descent follows a zig-zag path, often bouncing off the canyon walls (since the gradient is perpendicular to the surface), and we observe that it spends a lot of descending the canyon walls because they are the steepest feature. It indicates that gradient descent takes a larger number of steps to converge, and near the minima, it oscillates back and forth until it converges to the local minima. If the step size is large, it can overshoot and reach the opposite canyon wall on the next iteration, hence not the best search direction. The large positive eigenvalue of the Hessian corresponding to the eigenvector pointed in this direction indicates that this directional derivative is rapidly increasing, so an optimization algorithm based on the Hessian could predict the steepest direction is not actually a promising search direction in this context. Source: Ian Goodfellow \cite{b23}.}
    \label{fig:SGD}
    \vspace{-1.3em}
\end{figure}

\noindent In the case, where the loss surface is the shape of an elongated quadratic function or an ellipsoid as shown in Figure \ref{fig:SGD}, we can transform it into a spherical shape using a whitening transform since the Hessian inverse  $H^{-1}$ spheres out of the error surface locally, and we can now perform gradient descent in the new coordinate system.

\subsection{Quadratic Function Optimization}
\label{Quadratic Function Optimization}
\noindent For a locally quadratic function, by rescaling the gradients with the inverse of the Hessian, Newton's method jumps directly to the minimum and thus converges in a single step. If the function is nearly quadratic, then this is a very good estimate of the minimizer of $f$. Since $f$ is twice differentiable, the quadratic model of $f$ will be very accurate when $x$ is near the local minima. When $f$ is not quadratic but can be locally approximated as a positive definite quadratic, and the Newton's method is updated iteratively using eq. \ref{Newton method} with quadratic rate of convergence. The following result as stated in \cite{b30} asserts the rate of converge for the Newton's method and is given below.\\

\noindent THEOREM 1. \textit{Suppose that $f$ is twice differentiable and that the Hessian $\nabla^2 f(x)$ is Lipschitz continuous in a neighborhood of a solution $x^*$ at which the sufficient conditions are satisfied. Consider the iteration $x_{k+1}=x_k+p_k$, where $p_k$ is given by $-\nabla^2 f_k^{-1} \nabla f_k$. Then:\\
(i) if the starting point $x_0$ is sufficiently close to $x^*$, the sequence of iterates converges to $x^*$.\\
(ii) the rate of convergence of $\left\{x_k\right\}$ is quadratic.\\
(iii) the sequence of gradient norms $\left\{\left\|\nabla f_k\right\|\right\}$ converges quadratically to zero.}

\begin{algorithm}[hbt!]
\caption{Newton's Method with Hessian Modification}
\label{alg:Alg2}
$\textbf{Initialize:}$ Initial point $x_0$. \\
\For {$k=0,1,2, \ldots$}{
    Factorize the matrix $B_k=\nabla^2 f\left(x_k\right)+E_k$, where $E_k=0$ if $\nabla^2 f\left(x_k\right)$ is sufficiently positive definite \\ otherwise, $E_k$ is chosen to ensure that $B_k$ is sufficiently positive definite.\\
    Solve $B_k p_k=-\nabla f\left(x_k\right)$.\\
    $x_{k+1} \leftarrow x_k+\alpha_k p_k$}
\end{algorithm}
\subsection{Newton's Method}
\label{Newton Method Explaination}
\noindent In Newton’s optimization algorithm, to find the local minimum $ x^{*} $
with iteration sequence of $ x_{0} \rightarrow x_{1} \rightarrow x_{2} \rightarrow \ldots \ldots . . \rightarrow x_k $ the Hessian $ \nabla f^{2}\left(x_{k}\right) $ must be positive definite $\forall k$ iteration steps else the search direction might not correspond to an actual the descent direction i.e., $\nabla f_k^Tp_k < 0$, where $p_k$ is the Newton search direction. If $f$ is strongly convex then f converges quadratically fast to $x^* = \arg \min_{x} f(x)$ i.e.,
\begin{equation*}
\left\|x_{k+1}-x_*\right\| \leq \frac{1}{2}\left\|x_k-x_*\right\|^2, \quad \forall k \geq 0.
\end{equation*}
The update rule is given as follows:
\begin{equation*}
    \label{Newton method}
    \boldsymbol{\theta} \leftarrow \boldsymbol{\theta} - \boldsymbol{H^{-1}g}
\end{equation*}
\noindent where $H$ is the Hessian matrix of the loss function $L$ with respect to $\theta$. Since the Hessian ($H =  \nabla f^{2}(x_{k})$) is symmetric, we can determine the Newton update using Cholesky\footnote{Every symmetric positive definite matrix $A$ can be written as $A = LDL^T$ where $L$ is a lower triangular matrix with unit diagonal elements and $D$ is a diagonal matrix with positive elements on the diagonal} algorithm.  An essential feature of Newton's method is that it is independent of linear changes in coordinates. If $H$ is positive semi-definite, then there is nothing that confirms that the iterator has converged to a minimizer since the higher derivatives of $f(x)$ are unknown. The algorithm will diverge if the Hessian is not positive definite ($\lambda_i \leq 0$ indicates flat regions or directions of negative curvature).\\

\noindent Thus, alternative methods to ensure that eq. \ref{Newton method} corresponds to a descent direction while retaining the second-order information in $\nabla^2 f_k$ with superlinear rate of convergence is described in the following sections. In particular, the computation of the Hessian $H$ is computationally expensive and error-prone. For example, in Quasi-Newton methods an approximation to the inverse of the Hessian $B_k$ is computed at each step using a low-rank formula while a trust region approach, in which $\nabla^2 f_k$ is used to form a quadratic model that is minimized in a ball around the current iterate $x_k$.
\begin{figure}[hbt!]
\centering{\includegraphics[scale=0.4]{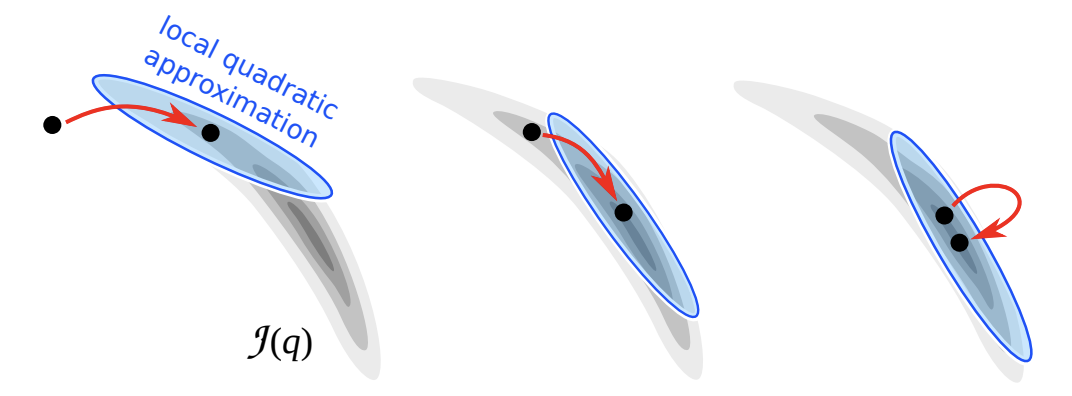}}
    \caption{Newton method constructs a local quadratic approximation of the function about the current point, moves towards the minimizer of this quadratic model, and then iteratively progress towards the global minima.  Source: Nicholas Vieau Alger \cite{b17}.}
    \label{fig:quadratic approximation}
    \vspace{-1.4em}
\end{figure}
\subsection{Dynamics of Optimization}
As discussed in Section \ref{Newton Method Explaination}, the major drawback of the Newton method is that the update step can result in the wrong direction if the eigenvalue $\lambda_i$ of the Hessian $H$ is negative i.e., it moves along the eigenvector in a direction opposite to the gradient descent step and thus moves in the direction of increasing error towards $\theta^{*}$. To address this issue, generalized trust region \cite{b30}, saddle-free Newton \cite{b5} and  natural gradient descent method are used and is discussed briefly as follows:
\begin{itemize}
    \item In the trust region method, a damping constant $\alpha$ is added along the diagonal of the Hessian to offset the direction of negative curvature, which is equivalent to  $\lambda_{i} + \alpha$. However, to ensure that this is a descent direction, one must ensure that $\lambda_{i} + \alpha > 0$, and can result in small step size along the eigen-directions. 
    \item Although, the truncated Newton method ignores directions of negative curvature it can get struck at saddle points. However, since the natural gradient descent relies on the Fisher information matrix $F$ to incorporate the curvature information of the parameter manifold where the update rule is given as:
    \begin{equation*}
        w_{t+1}=w_t-\eta F\left(w_t\right)^{-1} g_{t}
    \end{equation*}
    \item However, \cite{b5} argue that natural gradient descent method can also suffer from negative curvature and can converge to a non-stationary point when the Fisher matrix is rank-deficient. The saddle-free Newton method provides a elegant solution to this, where the update rule is given as:
    \begin{equation*}
        \Delta \theta=-\nabla f|\mathbf{H}|^{-1}
    \end{equation*}
    \item Since, it does not utilize the second-order Taylor-series approximation to leverage the information, as in classical methods, it can move further in direction of low curvature and escape saddle points as shown in Figure \ref{saddle free newton method}.
\end{itemize}

\subsection{Gauss-Newton and Levenberg-Marquardt algorithm}
Following \ref{Mathematical Preliminaries}, we define the non-linear least square consider a set of m  points $(x_{i}, y_{i})$ and the curve $y = f(x, \theta)$, where $\theta \in R^{n}$ denotes the model parameters and $i \in [m]$, with $m \geq n$. We define the non-linear least square problem as:
$$
\min_{\theta} \|y_{i} - f(x_{i}, \theta)\|_2^2 = \min_{\theta} \sum_{i=1}^m \| r_{i}\|_2^2
\vspace{0.3em}
$$
where $r_i = y_{i} - f(x_{i}, \theta) $ are the residuals $\forall i = 1, 2, ..., m$. The Gauss-Newton \cite{b10, b52} method utilizes the square of the Jacobian (not the Hessian) and is used to minimize non-linear least square problems. It can be viewed as a modified Newton's method with line search where we approximate the Hessian as $\bigtriangledown f_{k} \approx J_k^{T}J_{k}$ and has $O\left(N^{3}\right) $ complexity. The update rule is given as:
\begin{equation*}
    \Delta \theta = -(J_k^{T}J_{k})J_k^{T}r_{i}
    \vspace{0.4em}
\end{equation*}
\noindent where $J_{i}(x) = \left[\begin{array}{c}
\nabla r_1(x)^T,
\nabla r_2(x)^T, 
\hdots,
\nabla r_m(x)^T
\end{array}\right]^T$ is Jacobian of the residual. Note that, the Gauss-Newton method does not require the computation of the individual residual Hessians $\bigtriangledown^2 r_i$ and saves significant computational time. Also, near the minimum since the residuals are significantly smaller i.e., $\| r_{i} |\ \leq \epsilon$, this leads to rapid-convergence.\\

\noindent The Levenberg-Marquardt \cite{b53, b54} algorithm is an approximation to Gauss-Newton method that uses a diagonal approximation to the Hessian and takes care of extreme directions of curvature, thus preventing the update from moving in the wrong direction. This addresses the issue namely, when the Jacobian $J(x)$ is rank-deficient (i.e., the columns are not linearly independent) while maintaining similar local convergence properties since we only replace the line search with a trust-region method. The new update rule is given as:
$$
\Delta \theta = -(J_k^{T}J_{k} + \lambda I)J_k^{T}r_{i}
$$
\noindent where $\mu$ is the regularization parameter and $I$ is the Identity matrix. The regularization parameter is used to address the issue when the Hessian is not positive definite and has directions that correspond to small negative eigenvalues.
\begin{figure}[hbt!]
\centering
\begin{subfigure}
  \centering
  \includegraphics[width=.25\linewidth]{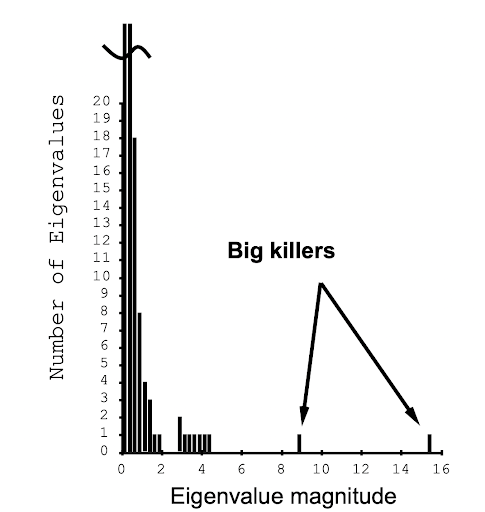}
  \label{fig:sub1}
\end{subfigure}%
\begin{subfigure}
  \centering
  \includegraphics[width=.25\linewidth]{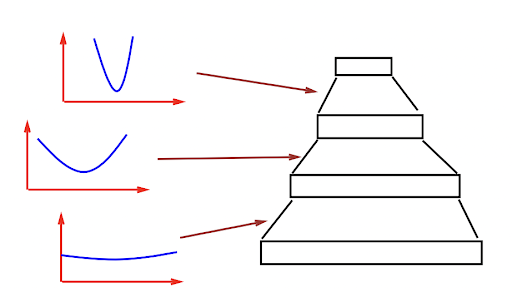}
  \label{fig:sub2}
\end{subfigure}
\caption{Eigenvalue spectrum in a 4 layer shared weights network. Source:Yann LeCun \cite{b1}}
\end{figure}
\subsection{Conjugate Gradients}
The conjugate gradient \cite {b55, b56} optimization is an iterative method that utilizes the conjugate directions i.e. orthogonal directions to perform a gradient update step and thus avoids the explicit computation of the inverse of the Hessian. This approach is used to overcome the weakness of the gradient descent method by aligning the current search direction along the current gradient step with a memory component given by the linear combination of previous directions. For a quadratic function, the gradient descent method fails to exploit the curvature information present in the Hessian matrix, wherein it descends down the canyon iteratively by following a zig-zag pattern and wastes most of its time by repeatedly hitting the canyon walls and makes no significant progress. This happens because each line search direction is orthogonal to the previous direction since the minimum of the objective along a given direction corresponds to the steepest descent direction at that point.\\

\noindent Thus, this descent direction does not preserve the progress in the previous direction and will undo the progress in that direction, and we need to reiterate to minimize the progress made in the previous direction. Thus, we obtain an ineffective zig-zag pattern of progress towards the optimum, and because of the large step size, it often overshoots the minima. The conjugate gradients address this problem by making the search directions orthogonal, i.e., the current search direction is conjugate to the previous line search direction, and it will not spoil the progress made in the previous iteration.\\

\noindent This method is only suitable for solving positive definite systems, i.e. when the Hessian has all positive eigenvalues. The orthogonality condition rightly finds the minimum along the direction $ d_{i} $ since the eigenvector $ e_{i+1} $ is orthogonal to $ d_{i} $. Two directions $d_{t-1}$ and $d_{t}$ are defined as conjugate if it satisfies:
$$
\boldsymbol{d}_{t}^{\top} \boldsymbol{H} \boldsymbol{d}_{t-1}=0
$$
\noindent where $H$ is the Hessian matrix. Thus, this method is an example of Gram-Schmidt ortho-normalization, converging in atmost $k$ line searches. At training iteration $t$, the current search direction $d_{t}$ is computed as discussed in Algorithm \ref{alg:Alg4}.
\begin{algorithm}[hbt!]
\caption{Conjugate Gradient Method}
\label{alg:Alg3}
$\textbf{Initialize:}$ Initial point $x_0$. $r_{0} = d_{0}$.  \\
\While {$r_{k} \neq 0$}{
    Set $r_{0} \leftarrow A x_{0} - b_{0}$. \\
    $\alpha_{i} \leftarrow \frac{r_{i}^T r_{i}}{d_{i}^T A d_{i}}$.\\
    $x_{i+1} \leftarrow x_{i}+\alpha_{i} d_{i}$. \\
    $r_{i+1} \leftarrow r_{i}-\alpha_{i} A d_{i}$. \\
    $\beta_{i+1} \leftarrow \frac{r_{i+1}^T r_{i+1}}{r_{i}^T r_{i}}$. \\
    $d_{i+1} \leftarrow r_{i+1}+\beta_{i+1} d_{i}$}
\end{algorithm}
\noindent The parameter $\beta_{t}$ controls how much of the previous direction should contribute to the current search direction. Since, this line search method utilizes the eigenvectors of H to choose $\beta_{t}$ which could be computationally expensive, we mainly use two popular methods \cite{b57, b58} for computing $\beta_{t}$ and they are as follows:\\

\noindent 1. The Fletcher-Reeves:
$$
\beta_{t}=\frac{\nabla_{\boldsymbol{\theta}} J\left(\boldsymbol{\theta}_{t}\right)^{\top} \nabla_{\boldsymbol{\theta}} J\left(\boldsymbol{\theta}_{t}\right)}{\nabla_{\boldsymbol{\theta}} J\left(\boldsymbol{\theta}_{t-1}\right)^{\top} \nabla_{\boldsymbol{\theta}} J\left(\boldsymbol{\theta}_{t-1}\right)}
$$
2. The Polak-Ribiere:
$$
\beta_{t}=\frac{\left(\nabla_{\theta} J\left(\theta_{t}\right)-\nabla_{\theta} J\left(\theta_{t-1}\right)\right)^{\top} \nabla_{\theta} J\left(\theta_{t}\right)}{\nabla_{\theta} J\left(\theta_{t-1}\right)^{\top} \nabla_{\theta} J\left(\theta_{t-1}\right)}
$$
\\

\noindent For non-linear conjugate gradient method, we set $\beta_{t}$ to zero i.e., it restarts at each step by forgetting the past search directions with the current direction given by the unaltered gradient at that point, to ensure that it is locally optimal and ensures faster convergence. Also, the initial point has a strong influence on the number of the steps taken for convergence, and it is observed that the Fletcher-Reeves method converges only if the initial point is sufficiently close to the desired minima and thus, the Polak-Ribiere method converges much more quickly.

\begin{figure}[hbt!]
\centering{\includegraphics[scale=0.3]{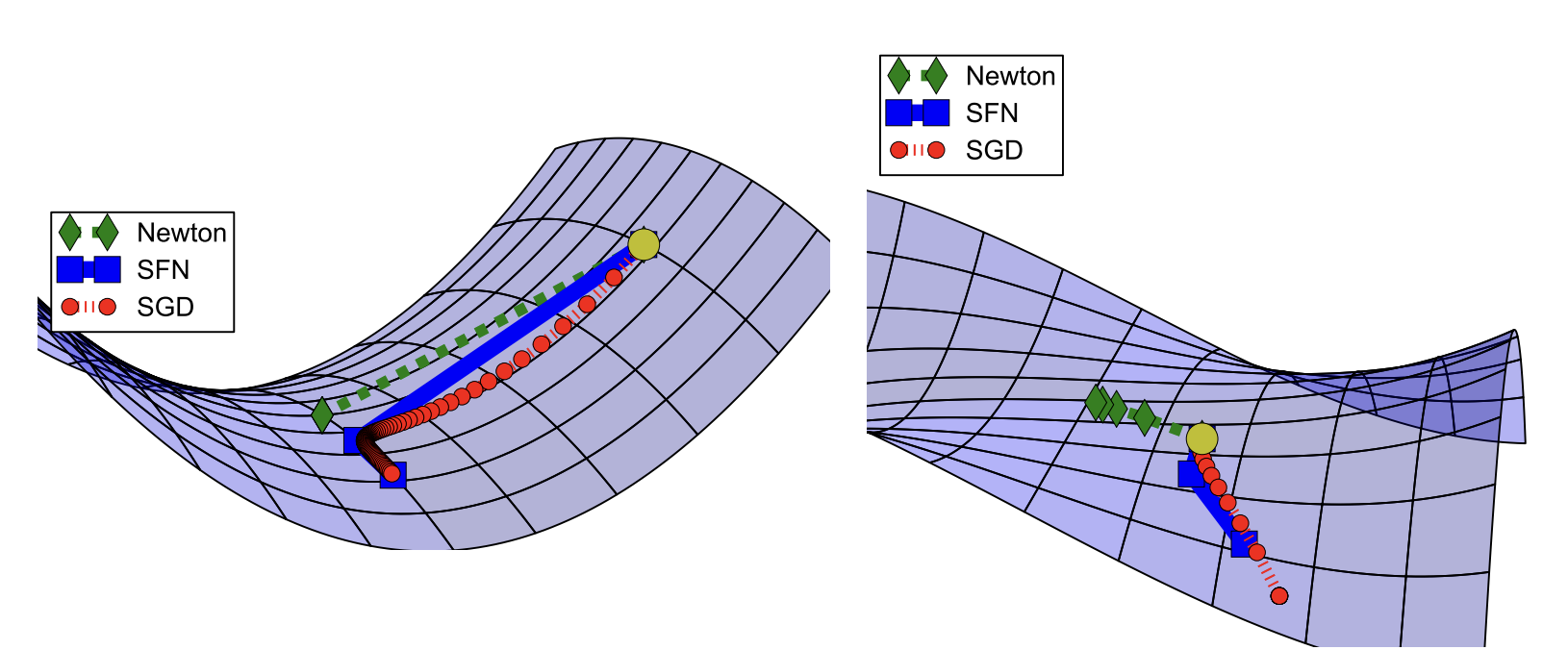}}
    \caption{Illustration of various optimization methods for various quadratic functions. The yellow dot indicates the starting point. Source Pascanu \cite{b5}. }
    \label{saddle free newton method}
\end{figure}

\subsection{BFGS}
The Quasi-Newton Broyden-Fletcher-Goldfarb-Shanno (BFGS) \cite{b59} method iteratively builds an approximation of the inverse of the Hessian at each step. The BFGS incorporates the second-order information using a low-rank approximation of the Hessian and is thus similar to the method of conjugate gradients without the computational burden. Here the explicit calculation of the Hessian is not required since we now use the secant equation to construct the approximation matrix by successively analyzing the gradient vector.
\begin{algorithm}[hbt!]
\caption{BFGS Method}
\label{alg:Alg4}
$\textbf{Initialize:}$ Initial point $x_0$, inverse Hessian approximation $B_{0}$, where $B_{0}$ is positive definite.\\
\For {$k=0,1,2, \ldots$}{
Determine the search direction $p_k$ by solving $B_k \mathbf{p}_k=-\nabla f\left(\mathbf{x}_k\right)$.\\
A line search is performed in this direction to determine the step size $ \varepsilon_k^{*}$, satisfying the Wolfe \\ conditions.\\
Define $s_k=x_{k+1}-x_k$ and $y_k=\nabla f(x_{k+1})-\nabla f(x_{k})$.\\
Apply Update $x_{k+1} \leftarrow x_{k} + s_{k}$.\\
$B_{k+1}=\left(I-\rho_k s_k y_k^T\right) B_k\left(I-\rho_k y_k s_k^T\right)+\rho_k s_k s_k^T$, where $\rho_{k}=\frac{1}{y_k^T s_k}$.}
\end{algorithm}
\noindent We perform a series of line searches along this direction, and since it takes more steps to converge due to the lack of precision in the approximation of the true inverse Hessian i.e., $H_k = B_k^{-1}$, as illustrated in Algorithm \ref{alg:Alg3}. The computational complexity for the BFGS method is a $ O\left(N^{2}\right) $ since matrix inversion is not required. The rank-two approximation in Alg \ref{alg:Alg3} can be considered as a diagonal and a low-rank approximation to the Hessian $H_k$. Since it does not utilize only second-order information, it is sometimes more efficient than the Newton's method for non-quadratic functions with super-linear rate of convergence and the cost per iteration is usually lower. Moreover, it is estimated that BFGS has self-correcting properties, i.e., when the previous estimate is bad and slows down the convergence, then BFGS will correct itself in the next few steps. We mainly compute the approximation to the Hessian matrix $H_{k}$ using: 1. Finite difference method 2. Hessian-vector products. 3. Diagonal Computation. 4. Square Jacobian approximation.
\begin{figure}[hbt!]
\centering
\begin{subfigure}
  \centering
  \includegraphics[width=.25\linewidth]{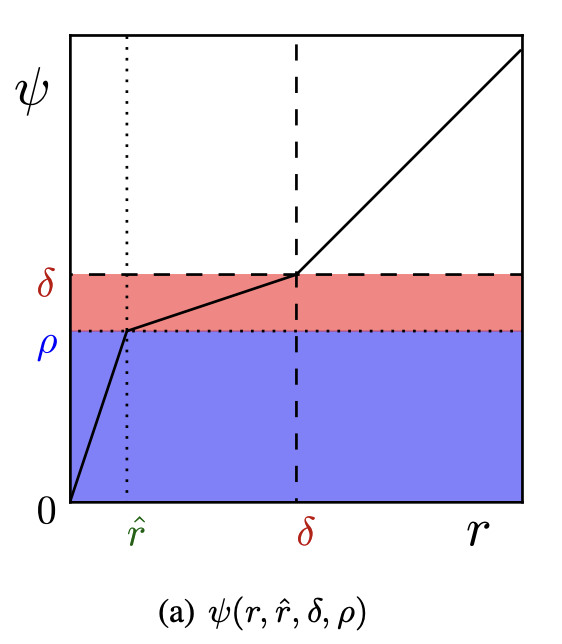}
  \label{fig:sub1}
\end{subfigure}%
\begin{subfigure}
  \centering
  \includegraphics[width=.25\linewidth]{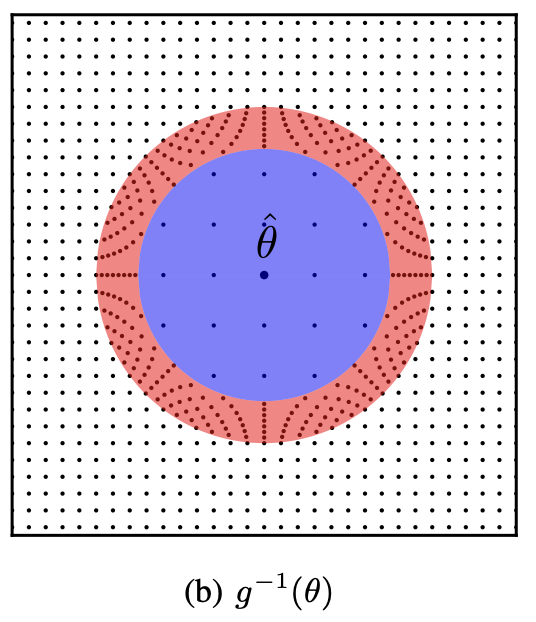}
  \label{fig:sub2}
\end{subfigure}

\caption{An example of a radial transformation on a 2-dimensional space to illustrate the effect of a local perturbation on the relative flatness between the minima. We can see that only the area in blue and red, i.e. inside the ball $B_2(\hat{\theta}, \delta)$, are affected. Here, $\psi(r, \hat{r}, \delta, \rho)= \mathbb{1}(r \notin[0, \delta]) r+\mathbb{1}(r \in[0, \hat{r}]) \rho \frac{r}{\hat{r}} +\mathbb{1}(r \in] \hat{r}, \delta])\left((\rho-\delta) \frac{r-\delta}{\hat{r}-\delta}+\delta\right)$  represents the function under consideration $g^{-1}(\theta)=\frac{\psi\left(\|\theta-\hat{\theta}\|_2, \hat{r}, \delta, \rho\right)}{\|\theta-\hat{\theta}\|_2}(\theta-\hat{\theta})+\hat{\theta}$ \$ is the radial-basis transformation respectively.  B. Source: Dinh \cite{b4}}
\end{figure}
\subsection{L-BFGS}
Since the memory costs of storing and manipulating the inverse of the Hessian is prohibitively large for deep neural nets, we utilize the L-BFGS \cite{b60} method to circumvent this issue. It maintains a simple and compact approximation of the Hessian matrices by constructing an Hessian approximation by using only the $m$ most recent iteration to incorporate the curvature information, where typically $3 < m < 20$. Thus, instead of storing the fully dense $n^{*}n$ approximators, we modify $H_{k}$ by implicitly storing only a certain number of correction pairs ($m$) and thus include the curvature information from only the $m$ recent iteration (we store only a set of $n$ length vectors). Thus, we add and delete information at each stage, discard the past $m$ correction pairs, and start the process with the new {$s_k, y_k$} pairs.\\ 

\noindent This ensures to an optimal rate of convergence and may also outperform Hessian-free methods, since we only maintain a fixed memory system that iteratively discards the curvature information from the past since it is not likely to influence the behavior of the Hessian in the current iteration, thus saving computational storage. The update rule is exactly similar to the BFGS method as discussed in Algorithm \ref{alg:Alg4}. The only modification is that we discard the vector pair $\left\{s_{k-m}, y_{k-m}\right\}$ from storage and the $s_k, y_k$ and $x_k$ updates follows from Algorithm \ref{alg:Alg4}. The main weakness of L-BFGS is that it converges slowly for functions, where the Hessian is ill-conditioned and contains a wide distribution of eigenvalues.\\


\subsection{ADAHESSIAN}
\begin{figure}[hbt!]
\centering
\begin{subfigure}
  \centering
  \includegraphics[width=.3\linewidth]{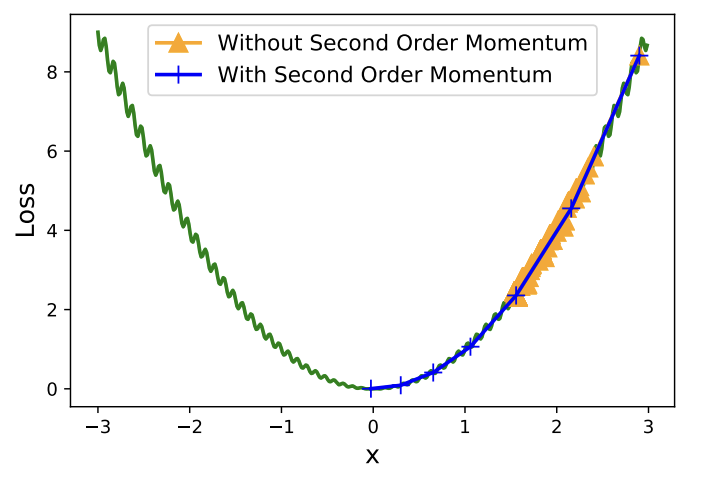}
  \label{fig:sub1}
\end{subfigure}%
\begin{subfigure}
  \centering
  \includegraphics[width=.3\linewidth]{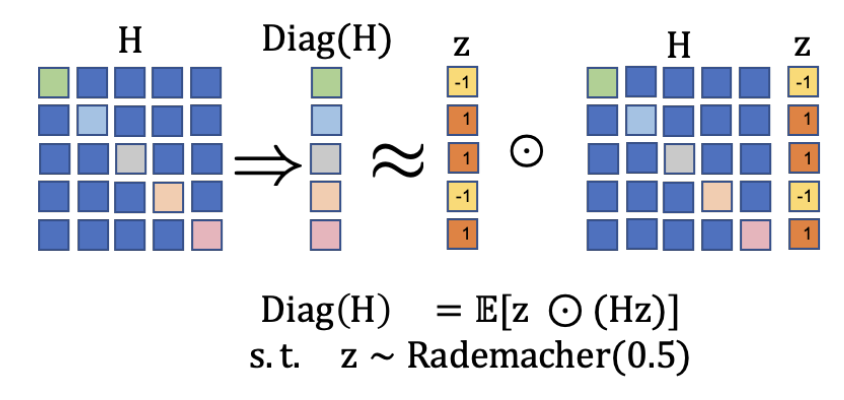}
  \label{fig:sub2}
\end{subfigure}
\caption{Illustration of the local and global curvature wherein the local curvature information is avoided using an exponential moving average. Adahessian converges in $1000$ iterations without a moving average, while it takes only seven iterations with the moving average enabled. Source: Zhewei Yao \cite{b34}}
\end{figure}
\noindent Adahessian \cite{b34} is a second-order stochastic approximation method that incorporates the curvature information present in the Hessian for faster convergence and reduced memory overhead using:\\

\noindent 1. Hessian diagonal approximation using Hutchinson's method.\\
\noindent  2. RMSE exponential moving average to reduce the variations in the diagonal approximation of the Hessian.\\
\noindent  3. Block diagonal averaging to reduce the variance of Hessian diagonal elements.\\

\noindent As discussed earlier, second-order methods involve preconditioning the gradient with the inverse of the Hessian, which is designed to accelerate learning for ill-conditioned problems i.e., flat curvatures in some directions and sharp curvature in other directions by rescaling the gradient vector along each of its directions. It handles extreme curvature directions by appropriate gradient rotation and scaling and ensures convergence to a critical point of any sort (local minima). It is thus more reliable than other adaptive first-order methods. But this comes at a prohibitive cost since it is computationally infeasible to compute the Hessian at each iteration and not scalable for modern neural network architectures. Thus, we approximate the Hessian matrix as a diagonal operator given as:
$$
\Delta w=\operatorname{diag}(\mathbf{H})^{-k} \mathbf{g}
$$
\noindent where $ \Delta w $ is the weight update and $ 0 \leq k \leq 1 $ is the Hessian power. The Hessian diagonal denoted as $D = \operatorname{diag}(\mathrm{H})$ is computed using Hutchinson’s method. To obtain the Hessian matrix $H$ without explicit computation, they utilize the Hessian-free approximation method given as:
$$
\frac{\partial \mathbf{g}^{T} z}{\partial \theta}=\frac{\partial \mathbf{g}^{T}}{\partial \theta} z+\mathbf{g}^{T} \frac{\partial z}{\partial \theta}=\frac{\partial \mathbf{g}^{T}}{\partial \theta} z=\mathbf{H} z
$$
\noindent where $z$ is a random vector with Rademacher distribution. The Hessian diagonal is given as the expectation of $ z \odot H z $ which corresponds to optimal performance on various tasks.
$$
\boldsymbol{D}=\operatorname{diag}(\mathbf{H})=\mathbb{E}[z \odot(\mathbf{H} z)]
$$
\noindent This has the computational overhead equivalent to a gradient backpropagation step. They employ a moving-average term (spatial averaging) and Hessian momentum (the update rule is similar to Adam) to smooth out stochastic variations in the curvature information per iteration. This leads to faster convergence for strict and smooth convex functions.
$$
D^{(s)}[i b+j]=\frac{\sum_{k=1}^{b} D[i b+k]}{b}, \text { for } 1 \leq j \leq b, 0 \leq i \leq \frac{d}{b}-1
$$
\noindent where $ D^{(s)} $ is spatially averaged Hessian diagonal and $b$ is the block size. We implement momentum for the Hessian diagonal $H$ with spatial averaging as shown in Table \ref{tab:equations}.
\begin{table}[htp!]
\centering
\begin{tabular}{|c|c|}
Term & Equation\\
$\overline{\boldsymbol{D}}_t$ & $\sqrt{\frac{\left(1-\beta_2\right) \sum_{i=1}^t \beta_2^{t-i} \boldsymbol{D}_i^{(s)} \boldsymbol{D}_i^{(s)}}{1-\beta_2^t}}$\\
$m_t$ & $\frac{\left(1-\beta_1\right) \sum_{i=1}^t \beta_1^{t-i} \mathbf{g}_i}{1-\beta_1^t}$\\
$v_t$ & $\left(\overline{\boldsymbol{D}}_t\right)^k$\\
\end{tabular}
\caption{Update Rule}
\label{tab:equations}
\vspace{-1em}
\end{table}
\noindent The parameter update rule is given as, 
$\theta_{t+1} = \theta_{t} - \eta m_{t}/v_{t}$, where $m_{t}$ and $v_{t}$ are the first and second order moments for AdaHessian with $0 < \beta_{1} < \beta_{2} < 1$.

\begin{figure}[htbp]
\begin{minipage}{0.5\linewidth}
\centering
\begin{tabular}{lcc}
\hline Optimizer & CIFAR-10 & CIFAR-100 \\
\hline SGD & $94.33 \pm 0.1$ & $79.12 \pm 0.54$ \\
Polyak & $95.62 \pm 0.07$ & $78.11 \pm 0.37$ \\
Adam & $94.76 \pm 0.23$ & $77.24 \pm 0.12$ \\
\hline
LOOKAHEAD & $\textbf{95.63} \pm \textbf{0.03}$ & $\textbf{79.42} \pm \textbf{0.05}$ \\
\hline
\end{tabular}
\end{minipage}%
\begin{minipage}{0.5\linewidth}
\centering
\begin{tabular}{lccc}
\hline Dataset & \multicolumn{2}{c}{Cifar10} & ImageNet \\
& ResNet20 & ResNet32 & ResNet18 \\
\hline SGD & 93.01 $\pm$ 0.05 & $\mathbf{94.23} \pm \mathbf{0.17}$ & 70.01 \\
Adam & 90.73 $\pm$ 0.23 & 91.53 $\pm$ 0.1 & 65.42 \\
AdamW & 91.96 $\pm$ 0.15 & 92.82 $\pm$ 0.1 & 67.82 \\
\hline ADAHESSIAN & $\mathbf{93.11} \pm \mathbf{0.25}$ & 93.08 $\pm$ 0.10 & $\mathbf{70.12}$ \\
\hline
\end{tabular}
\end{minipage}
\caption{(left) We report the test accuracy (+ standard deviation) across $10$ trails using different optimizers on the CIFAR-$10/100$ dataset using the ResNet$18$ and architecture. (right) We report the model accuracy (+ standard deviation) across $10$ trails using different optimizers on the CIFAR-$10/100$ and ImageNet dataset using the ResNet$20$ and ResNet$32$ architecture. We train all our models for $200$ epochs with a batch size of $128$ and an initial learning rate of $0.1/0.001/0.01$ for SGD/Adam/AdamW respectively. For Adahessian we set block size equal to $9$, $k=1$ and a learning rate of $0.15$; while for Lookahead we set $k = 5$ and use a learning rate of $0.1$. The exact training details as mentioned in \cite{b34, b61} was used for model training.}
\end{figure}


\subsection{Newton-CG method}
The line search Newton-CG \cite{b30} method utilizes the Hessian-vector products ($\nabla^{2} f_{k}$) has attractive global convergence properties with super-linear convergence, and results in optimal search directions when the Hessian is indefinite. This Hessian-free (HF) method does not require explicit knowledge of the Hessian but uses the finite-difference method to access the actual Hessian at each point. Here, we compute the product of the Hessian ($ \nabla^{2} f_{k} $) and the vector $d$ using a finite-differencing technique to get the approximation:
$$
\nabla^{2} f_{k} d \approx \frac{\nabla f\left(x_{k}+h d\right)-\nabla f\left(x_{k}\right)}{h}
$$
\noindent for some small differencing interval h. Also, using the Gauss-Newton matrix instead of the Hessian results in better search directions since $G$ is always positive definite and thus avoids the problem of negative curvature. CG is designed for positive-definite systems, and the Hessian is indefinite for points far away from the solution; we can terminate early or use negative directions when they are detected to account for this. Since the HF approach has no access to the true Hessian, we may run across directions of extremely low curvatures or even negative curvatures. Thus, the search directions can only result in small function reductions even after many function evaluations in line search. If we are not careful, such a direction could result in a very large CG step, possibly taking the outside of the region where the Newton approximation is even valid. One solution is to add a damping parameter to the Hessian given as $ \mathrm{B}=\mathrm{H}+\lambda \mathrm{I} $, where $I$ is the Identity matrix. This reconditions $H$ and controls the length of each CG step, mimicking a trust region method.
$$
\rho_{k}=\frac{f\left(x_{k}+p_{k}\right)-f\left(x_{k}\right)}{q_{k}\left(p_{k}\right)-q_{k}(0)} .
$$
$$
\text { If } \rho_{k}<\frac{1}{4}, \text { then } \lambda \leftarrow \alpha \lambda. \hspace{0.4em}
\text { If}  \hspace{0.5em} \rho_{k}>\frac{3}{4}, \text { then } \lambda \leftarrow \alpha^{-1} \lambda .
$$
The main drawback of this method is when $\nabla f(x_k)$ is nearly singular, and thus requiring many function evaluations without significant reduction in the function values. To alleviate this difficulty, trust-region Newton-CG method and Newton-Lanczos method can be used respectively. 


\subsection{Lookahead}
Lookahead \cite{b61} optimizer is orthogonal to the previous approaches and proposes a distinct way for weight updates in which it maintains two sets of weights namely, slow weights ($\phi$) and fast weights ($\theta)$. It maintains an inner loop of optimization for $k$ steps using standard methods such as the SGD or Adam for iterative fast weights update and follows it up with a linear interpolation of the slow weights along the direction given as $ \theta-\phi $ in the weight space. Thus, after each slow weights update, we reset it as the new fast weights, making rapid progress in low curvature directions with reduced oscillations and quicker convergence. The slow weights trajectory is characterized as an exponential moving average i.e., Polyak averaging has strong theoretical guarantees of the fast weights. After $k$ inner-loop steps, the slow weights is updated as follows:
$$
\begin{aligned}
\phi_{t+1} &=\phi_{t}+\alpha\left(\theta_{t, k}-\phi_{t}\right) \\
&=\alpha\left[\theta_{t, k}+(1-\alpha) \theta_{t-1, k}+\ldots+(1-\alpha)^{t-1} \theta_{0, k}\right]+(1-\alpha)^{t} \phi_{0}
\end{aligned}
$$
\noindent where $\alpha$ is the learning rate for slow weights $\phi$. For a choice of optimizer $A$ and a mini-batch of $d$ data points the update rule for fast weights is:
$$
\theta_{t, i+1}=\theta_{t, i}+A\left(L, \theta_{t, i-1}, d\right)
$$

\section{Conclusion}
We summarize first-order and second-order optimization methods from a theoretical viewpoint using the optimization literature. Since high error, local minima traps do not appear when the model is overparameterized, and local minima with high error relative to the global minimum occur with an exponentially small probability in $N$. Critical points with high cost are far more likely to be saddle points. Also, since second-order methods are computationally intensive, the computational complexity of computing the Hessian is $O(N^3)$, and thus practically only networks with a small number of parameters can be trained via Newton's method.

\section{Acknowledgments}
We thank the anonymous reviewers for their review and suggestions.
\bibliographystyle{ACM-Reference-Format}

\vspace{12pt}
\end{document}